\documentclass{article}

\usepackage{microtype}
\usepackage{graphicx}
\usepackage{booktabs} 

\usepackage{hyperref}

\usepackage{nicefrac}

\usepackage[accepted]{icml2025}

\usepackage{amsmath}
\usepackage{amssymb}
\usepackage{mathtools}
\usepackage{amsthm}

\usepackage[capitalize,noabbrev]{cleveref}

\theoremstyle{plain}

\theoremstyle{definition}

\theoremstyle{remark}

\usepackage[textsize=tiny]{todonotes}
\usepackage{tabularx}
\usepackage{algorithm}
\usepackage{algorithmic}
\usepackage{url}
\usepackage{multirow}
\usepackage{comment}
\usepackage{subcaption}
\usepackage{xcolor}
\usepackage{amsmath,amsfonts,bm}
\usepackage{tikz}
\usetikzlibrary{shapes.geometric, positioning, arrows.meta}

\def\eqref#1{equation~\ref{#1}}

\def\1{\bm{1}}

\def\rvw{{\mathbf{w}}}

\def\vtheta{{\bm{\theta}}}

\def\mK{{\bm{K}}}

\def\mQ{{\bm{Q}}}

\def\mV{{\bm{V}}}
\def\mW{{\bm{W}}}
\def\mX{{\bm{X}}}

\DeclareMathAlphabet{\mathsfit}{\encodingdefault}{\sfdefault}{m}{sl}
\SetMathAlphabet{\mathsfit}{bold}{\encodingdefault}{\sfdefault}{bx}{n}

\def\sG{{\mathbb{G}}}
\def\sH{{\mathbb{H}}}

\def\sS{{\mathbb{S}}}

\newcommand{\KL}{D_{\mathrm{KL}}}

\DeclareMathOperator*{\argmax}{arg\,max}
\DeclareMathOperator*{\argmin}{arg\,min}

\usepackage{graphicx}
\definecolor{highlightorange}{rgb}{1, 0.85, 0.4}
\definecolor{highlightred}{rgb}{1, 0.4, 0.4}
\newcommand{\fullwidthbox}[1]{%
    \noindent
    \colorbox{highlightorange}{\parbox{\dimexpr\linewidth-2\fboxsep}{#1}}%
}

\usepackage{stfloats}
\usepackage{ulem} 
\usepackage{subcaption}
\usepackage{wrapfig}
\renewcommand{\emph}[1]{\textit{#1}}

\icmltitlerunning{Compressing Large Language Models with Automated Sub-Network Search}

\begin{document}

\twocolumn[
\icmltitle{Compressing Large Language Models with Automated Sub-Network Search}




\begin{icmlauthorlist}
\icmlauthor{Rhea Sanjay Sukthanker}{freiburg}
\icmlauthor{Benedikt Staffler}{bosch}
\icmlauthor{Frank Hutter}{tubingen,freiburg}
\icmlauthor{Aaron Klein}{scadsai}

\end{icmlauthorlist}

\icmlaffiliation{freiburg}{University of Freiburg}
\icmlaffiliation{tubingen}{ELLIS Institute Tübingen}
\icmlaffiliation{bosch}{Bosch Center for Artificial Intelligence, Germany}
\icmlaffiliation{scadsai}{ScaDS.AI}

\icmlcorrespondingauthor{Rhea Sanjay Sukthanker}{sukthank@cs.uni-freiburg.de}

\icmlkeywords{Machine Learning, ICML}

\vskip 0.3in
]



\printAffiliationsAndNotice{{}} 

\begin{abstract}
Large Language Models (LLMs) demonstrate exceptional reasoning abilities, enabling strong generalization across diverse tasks such as commonsense reasoning and instruction following.
However, as LLMs scale, inference costs become increasingly prohibitive, accumulating significantly over their life cycle. 
In this paper we consider model compression for LLMs to reduce model size while improving downstream task performance.
We phrase this as a neural architecture search problem that automatically prunes structural components, such as attention heads, neurons, and layers by searching for the Pareto-optimal set of sub-networks balancing between performance and on-device latency. Compared to state-of-the-art structural pruning approaches and fine-tuned smaller sub-networks extracted from the pre-trained model, our method achieves upto $9.85\%$ improvement on average on 11 diverse downstream tasks, while achieving up to $22\%$ improvement of on-device latency.

\end{abstract}
\section{Introduction}

Large Language Models (LLMs) mark a significant breakthrough in artificial intelligence, powering applications such as chatbots, virtual assistants, and code generation tools. Despite their impressive capabilities, the sheer size of these models introduces considerable challenges. High inference costs, substantial memory footprints, and elevated latencies often hinder deployment in real-time or resource-constrained environments, such as mobile devices and embedded systems. Additionally, deploying large models at scale incurs significant operational expenses, limiting their widespread adoption. 

To mitigate these issues, LLM providers typical offer models in varying sizes. For example, LLaMA 2 is available with 7B, 13B, 34B, and 70B parameters~\citep{touvron2023llama}, allowing users to select a model that balances performance with computational cost. However, even smaller variants remain expensive to train. While the LLaMA 2 70B model required up to 1,720,320 GPU hours, the 13B model still demanded 368,640 GPU hours, highlighting the intensive resource demands across all scales.  

An alternative to training smaller models from scratch is compressing existing large models while preserving their performance.
Distillation~\citep{hinton-arxiv15} transfers knowledge to a smaller student model by aligning its predictions with those of a larger teacher model.
While distillation often results in better performance compared to training the same model with standard cross-entropy loss, it still demands a significant amount of computational resources and training data.
Furthermore, determining the optimal size of the student model is highly dependent on the downstream task and it often requires multiple trials to identify the appropriate student network.

Structural pruning~\citep{frantar-arxiv23,ashkboos-iclr24} offers an alternative by systematically removing redundant components, such as attention heads and neurons, to reduce model size and latency.
Unlike distillation, most structural pruning techniques require little to no additional adaptation, or at most a small fine-tuning step, making them significantly more cost-effective than training a model from scratch, whether via pre-training or distillation.
However, a key challenge remains: \textit{To what extent can a model be compressed without incurring significant degradation in performance?}

\begin{figure}[t]
    \centering
    \includegraphics[width=.49\linewidth]{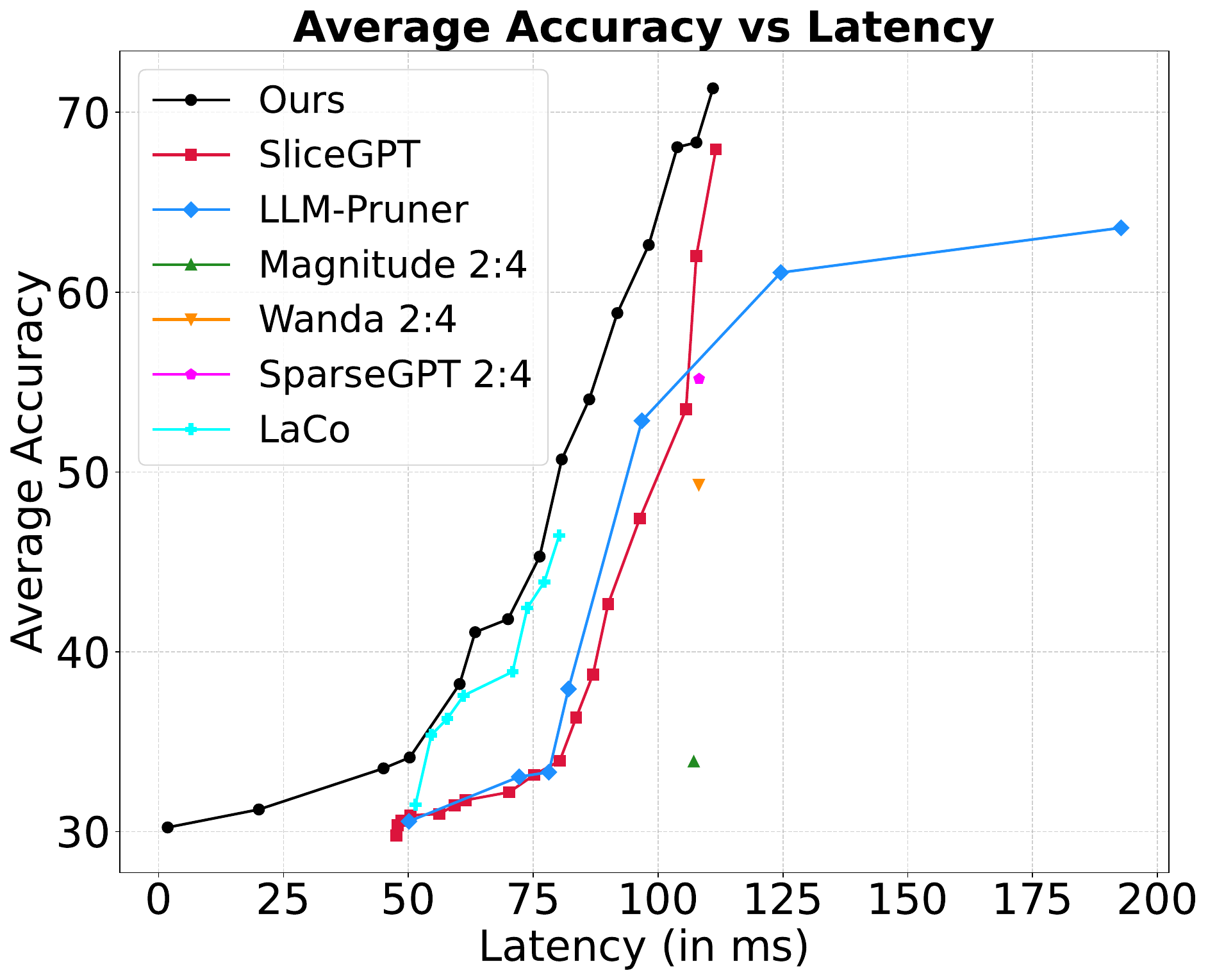}
    \includegraphics[width=.49\linewidth]{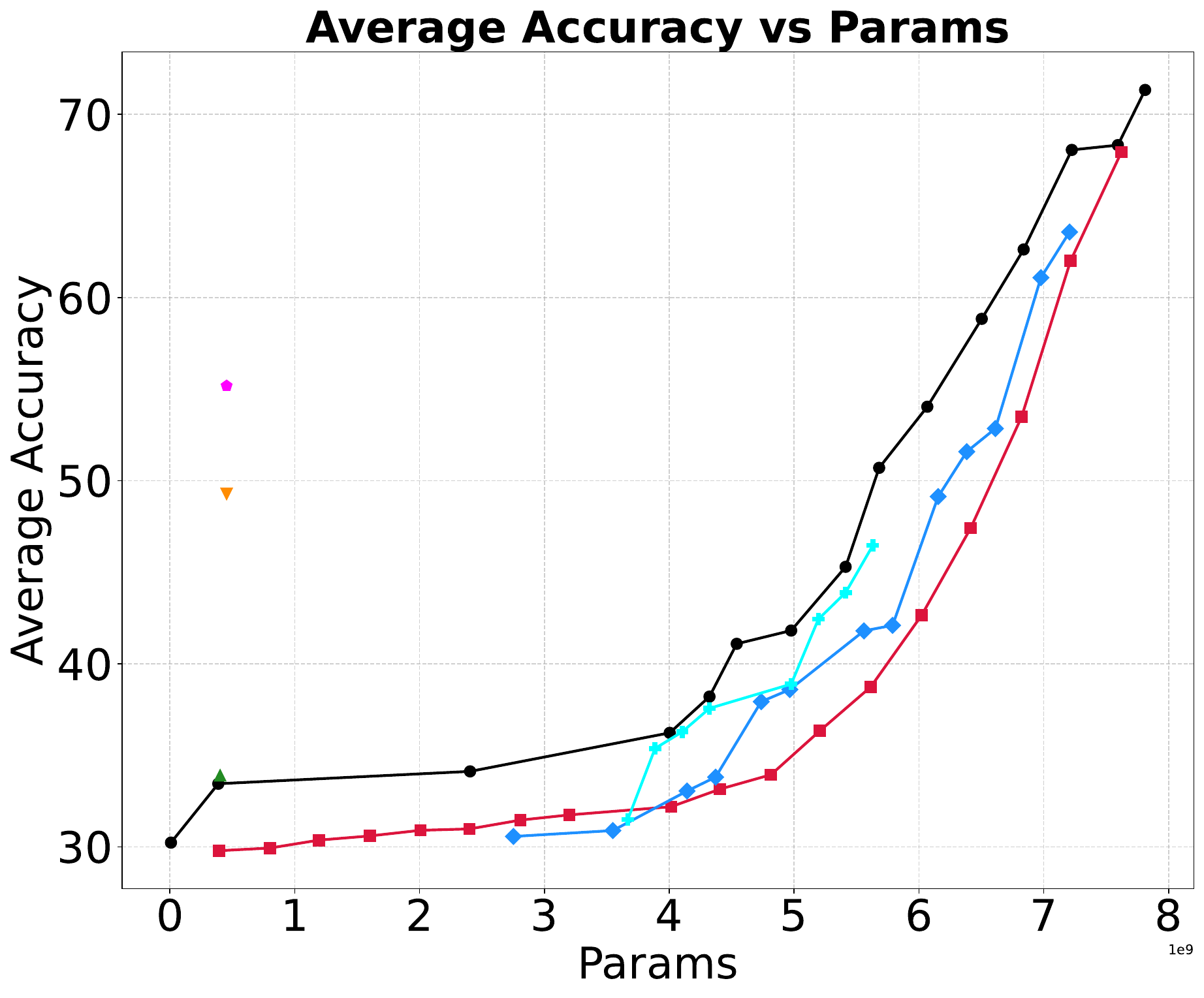}
    \caption{Comparison of Average Accuracy (on commonsense reasoning tasks) v/s Latency and Parameter Pareto-Fronts for different pruning methods on Llama-3.1-8B}
    \label{fig:combined_plot_llama_3.1_8b}
    \vspace{-5mm}
\end{figure}


In this work, we propose an automated structural pruning strategies that finds a Pareto-optimal set of sub-networks within a single run. 
We treat model compression as a neural architecture search (NAS)~\citep{elsken-jmlr2019} problem, where we treat the pre-trained network as super-network~\citep{yu-eccv20} and search for sparse sub-networks that optimally balance downstream performance and computational efficiency~\citep{klein-tmlr24,muralidharan2024compact}. Given that we can induce a search space on any pre-trained language model, this procedure is agnostic to the underlying language model architecture. While NAS has demonstrated success in compressing smaller encoder-only transformer models~\citep{klein-tmlr24}, scaling it to larger decoder-only models remains a significant challenge.
In this work, we discuss these limitations and introduce key extensions that enable scalable and efficient compression on LLM via NAS. The primary contributions of our work are:
\vspace{-2mm}
\begin{itemize}
\vspace{-2mm}
\item Applying NAS methods for compression requires fine-tuning a super-network to adapt sub-networks for pruning. However, the number of possible sub-networks grows exponentially with model size, posing significant challenges for larger models. We introduce improved sub-network selection strategies in Section~\ref{sec:method}, including \textbf{magnitude-based calibration} based on \textbf{importance sorted weights}, that expedite fine-tuning and enhance performance on downstream tasks.
\vspace{-2mm}
\item Fine-tuning LLMs with over a billion parameters typically requires substantial compute resources. Building on prior work~\citep{munoz2024lonas}, we empirically investigate the \textbf{integration of NAS with advanced parameter-efficient fine-tuning methods}, such as LoRA~\citep{hu-arxiv21}, enabling \textbf{scalable pruning} on consumer-grade GPUs.
\vspace{-2mm}
\item We demonstrate that our approach yields \textbf{a Pareto-optimal set of sub-networks with diverse latency-performance trade-offs}, as shown in Figure~\ref{fig:combined_plot_llama_3.1_8b}. These sub-networks consistently outperform structurally pruned models across benchmarks for commonsense reasoning and mathematical tasks, achieving superior downstream performance with significant latency gains.

\end{itemize}
\vspace{-4mm}
We structure the paper as follows: Section~\ref{sec:related} reviews related work on LLM compression. Section~\ref{sec:background} introduces the key concepts of the transformers and NAS, while Section~\ref{sec:method} delineates our approach for scaling NAS to LLMs. Section~\ref{sec:experiments} provides comprehensive empirical evaluation and ablations comparing our method to structural pruning baselines and analyzing its effectiveness. To facilitate reproducibility, we provide our code at \url{https://github.com/automl/automated-sub-network-selection}.
\section{Related Work}\label{sec:related}

\textbf{Pruning} reduces the size of trained neural networks by removing neurons or weights while preserving its predictive performance.
We distinguish between \textit{unstructured pruning}, which eliminates individual weights~\citep{sun-iclr2024,han2015deep,frantar2023sparsegpt,lualphapruning}, and \textit{structured pruning}
~\citep{ashkboos-iclr24,ma2023llm,yang2024laco}, which removes structural components, such as layers or attention heads.  

A prevalent unstructured pruning technique is weight magnitude pruning, which masks individual weights based on their magnitude~\citep{han2015deep}.
Alternative approaches use small calibration sets~\citep{sun-iclr2024}, iterative updates~\citep{frantar2023sparsegpt} or trainability of weights~\citep{lualphapruning} to identify and prune redundant weights.
In contrast, structured pruning targets larger components of the model, such as embedding dimensions~\citep{ashkboos-iclr24}, attention heads~\citep{he2024matters,michel2019sixteen}, or entire layers
~\citep{sajjad2023effect,men2024shortgpt,yang2024laco}.
While unstructured pruning often preserves performance at high sparsity levels, it results in sparse weight matrices that do not directly translate to latency improvements during inference.
Semi-structured N:M pruning~\citep{zhoulearning} bridges this gap by combining the benefits of both approaches, enabling latency gains (e.g., 2:4 sparsity on NVIDIA Ampere GPUs).  

Here we focus on structured pruning to identify optimal sparsity patterns that deliver measurable inference speedups.
Unlike previous structured pruning approaches that rely on handcrafted heuristics or scoring techniques, our objective is to automate the pipeline for generating pruned architectures, simplifying the process of creating efficient models.
Additionally, considering the growing KV cache sizes during inference in LLMs~\citep{pope2023efficiently}, we direct our efforts toward pruning modern architectures such as Llama-3.1~\citep{llama3-arxiv24}, which employ Grouped-Query-Attention (GQA) to mitigate the growing KV cache size, which is in contrast to prior works~\citep{xiasheared,an2024fluctuation}.

\textbf{Neural Architecture Search (NAS)} automates the design and optimization of neural networks by jointly optimizing objectives such as hardware efficiency and predictive performance~\citep{elsken-jmlr2019, white-arxiv23}. Two-stage NAS~\citep{yu-eccv20} trains a single super-network that contains a finite set of sub-networks, effectively acting as a structured pruning mechanism by identifying sparse, high-performing sub-networks that balance efficiency and accuracy.  

\citet{klein-tmlr24} demonstrated that two-stage NAS can outperform structural pruning methods on small encoder models.
Recent work~\citep{caiflextron,muralidharan2024compact} applied NAS to design Pareto-optimal, low-latency decoder-only architectures by learning routing mechanisms in LLaMA-2-7B.
Similarly, Minitron~\citep{muralidharan2024compact} utilized NAS, importance computation, and knowledge distillation to develop efficient versions of LLaMA-3.1-8B and Nemotron-4 15B. 
However, their fine-tuning pipelines and datasets are not publicly accessible, making a direct comparison challenging. 
Consequently, we compare only against established structured pruning methods.
Also, Minitron relies on multiple computationally expensive full fine-tuning runs to obtain a Pareto set of pruned networks, whereas our approach provides a full Pareto set in a cost-effective manner using a single parameter-efficient-fine-tuning~\citep{pfeiffer2023modular} run.  

\ifx\skiptodos\undefined
\fi

\vspace{-2mm}
\section{Background and Notations}\label{sec:background}

In this section we define the notations and key terminologies used throughout the paper. Section~\ref{subsec:background_transformer} describes the different components of decoder-only transformers, and Section~\ref{subsec:background_nas} discusses the foundational building blocks of NAS. 
\vspace{-4mm}
\subsection{Transformer Architecture}
\label{subsec:background_transformer}
The transformer architecture~\citep{vaswani-nips17} underpins many state-of-the-art LLMs. In this work, we focus on decoder-only architectures~\citep{radford-article19}, which are prevalent in leading open-source LLMs such as Llama~\citep{llama3-arxiv24}, Phi~\citep{phi3} and Gemma~\citep{team2024gemma} models.  

A transformer consists of an embedding layer that maps tokens to learnable vectors, followed by $L$ transformer blocks. Each block contains a self-attention mechanism and a fully connected feed-forward layer, with layer normalization~\citep{ba-arxiv16} or root-mean-square normalization~\citep{zhang-neurips19} applied before each of these layers. To maintain clarity in the notation, we will omit the layer index associated with each weight in the transformer block. The transformer blocks are followed by a language model head (a linear projection layer) that predicts the probabilities of the next tokens in the vocabulary space.

\textbf{Embedding Layer.} Each input token is encoded as a feature vector in $\mathbb{R}^{d_{model}}$ using an embedding matrix $\mW^{emb} \in \mathbb{R}^{V \times d_{model}}$, where $V$ is the vocabulary size and $d_{model}$ is the embedding dimension. This process generates an encoded input sequence $\mX \in \mathbb{R}^{T \times d_{model}}$ of length $T$.

%
\textbf{Multi-head self-attention.} (MHA), which captures cross-token dependencies, consists of $H$ attention heads $h_i, i \in [1, H]$ which perform attention over queries, keys and values $\mQ_i,\mK_i,\mV_i\in\mathbb{R}^{T \times d_{head}}$ calculated from the input $\mX$ to the MHA operation. Note however, that in principle $\mV$ can have different $d_{head}$ compared to $\mQ$ and $\mK$. 
More specifically,
\[
\mQ_i = \mX \mW^Q_i, \mK_i=\mX \mW^K_i, \mV_i=\mX \mW^V_i,
\]
where $\mW^Q_i, \mW^K_i, \mW^V_i \in \mathbb{R}^{d_{model} \times d_{head}}$.
The queries, keys and values for all heads can be calculated using a single linear layer with weights $\mW^{attn}\in\mathbb{R}^{d_{model} \times H \times 3 \times d_{head}}$.
A single attention head $i$ then computes
$X_i = \text{Att}(\mQ_i,\mK_i,\mV_i)$,
where $Att(\mQ, \mK, \mV) = Softmax\left(\frac{\mQ \mK^T}{\sqrt{d_{head}}}\right)\mV$ and the softmax is applied row-wise.
The outputs of all heads are concatenated and linearly transformed $    X = \text{concat}(X_1, \ldots, X_{H}) \cdot \mW^{proj}$ to produce the final output $X$, where $\mW^{proj} \in \mathbb{R}^{H \times d_{head} \times d_{model}}$.\\
\textbf{Grouped-Query Attention (GQA).} To reduce parameter overhead and enable efficient KV caching~\citep{pope2023efficiently}, GQA~\citep{ainslie-emnlp23} partitions the attention heads into $G$ groups. Within each group $g \in [1, G]$, all heads share a single key and value matrix, $\mK_{g}, \mV_{g} \in \mathbb{R}^{T \times d_{head}}$.
The attention weight matrix is now:
\[
\mW^{attn} \in \mathbb{R}^{d_{model} \times (H+2G)\times d_{head}}
\]

\textbf{Feed-Forward-Network.} The feedforward network (FFN), in a transformer block, which is responsible for element-wise processing of tokens, is defined as:  
\[
FFN(\mX) = \sigma(\mX \mW_{1} ) \mW_{2} 
\]
where $\mW_{1} \in \mathbb{R}^{d_{model} \times U}$ and $\mW_{2} \in \mathbb{R}^{U \times d_{model} }$, with expansion factor $U = r \cdot d_{model}$. The activation function $\sigma(\cdot)$ is typically GeLU~\citep{hendrycks-arxiv16}.

\textbf{Gated Linear Units (GLU).} A common variant of the FFN, used in models such as LLaMA~\citep{dubey2024llama}, is the Gated Linear Unit (GLU)~\citep{shazeer2020glu}. GLUs apply element-wise gating to the hidden layer, defined as:  
\[
FFN(\mX) =  \sigma(\text{sigmoid}(\mX\mW_{2} ) \odot (\mX\mW_{0})) \mW_{1},
\]
where $\mW_{0}, \mW_{2} \in \mathbb{R}^{d_{model} \times U}$ and $\mW_{1} \in \mathbb{R}^{U \times d_{model}}$ and $\odot$ denotes pointwise multiplication.
This formulation enhances expressivity by modulating the input before the linear transformation.  
\\
\textbf{RMSNorm.} Modern LLMs, typically replace the LayerNorm, which shifts and scales the activations, with RMSNorm, which only rescales the activations and hence is computationally simpler than LayerNorm. RMSNorm which is placed before GQA, GLU blocks and before the language model prediction head is defined, for every token $i$, as follows:
\[
\text{RMSNorm}(\mX_i) = \frac{\mX_i}{\sqrt{\frac{1}{d_{model}} \sum_{j=1}^{d_{model}} \mX_{ij}^2}} \odot \boldsymbol{\gamma}
\]
\textbf{Language Model Prediction Head.} Transformers use a linear layer with $\mW^{lm\_head} \in \mathbb{R}^{d_{model}\times V}$ as prediction head to output the logits in the vocabulary space.

\subsection{Two-Stage Neural Architecture Search}\label{subsec:background_nas}

Given a search space $\Theta$ composed of architectural design choices—such as the number of attention heads $H$ or transformer blocks $L$ —neural architecture search (NAS) seeks to identify the optimal architecture $\vtheta^{\star} = \argmin_{\vtheta \in \Theta} f(\vtheta)$ that minimizes an objective function $f(\vtheta)$, typically representing the validation error~\citep{elsken-jmlr2019}. This objective can be expressed as $f(\vtheta) = \mathcal{L}_{valid}(\vtheta, \rvw_{\vtheta}^{\star})$, where the network is trained to convergence $\rvw_{\vtheta}^{\star} = \argmin \mathcal{L}_{train}(\rvw_{\vtheta})$ where we encapsulate all trainable weights $\rvw_{\vtheta} \in \mathbb{R}^n$ in a single vector. Although the search space $\Theta$ is large, it is generally finite. NAS can be  extended to optimize multiple objectives $\min_{\vtheta \in \Theta} \left\{f_0(\vtheta), \dots, f_k(\vtheta)\right\}$, where $f_0, ..., f_k$ may include validation error, inference latency or parameter count.  

\ifx\skiptodos\undefined
\fi

Original evolutionary algorithms~\citep{real-aaai19} and reinforcement learning methods~\citep{zoph-iclr17a} have been proposed to tackle NAS.
However, evaluating $f(\vtheta)$ requires training and validating a neural networks for each $\theta$, resulting in significant computational overhead, making these methods impractical for large-scale models.  

\textit{Two-stage NAS}~\citep{yu-eccv20} mitigates this by training a single \textit{super-network} $\rvw_{\Theta}^{\star} = \argmin \mathcal{L}_{train}(\rvw_{\Theta})$ using a set of shared weights $\rvw_{\Theta} \in \mathbb{R}^d$. The super-network is constructed to encompass all possible architectures $\vtheta \in \Theta$ within the search space. Following super-network training, the validation error of a specific architecture $\vtheta$ is computed as $f(\vtheta) = \mathcal{L}_{valid}(\vtheta, \hat{\rvw}_{\vtheta})$, where $\hat{\rvw}_{\vtheta} \subseteq \rvw_{\Theta}$ represents a subset of the super-network's weights. This enables efficient evaluation without requiring independent training for each architecture, reducing computational costs by orders of magnitude. We refer to architectures $\vtheta$ that utilize subsets of the super-network weights as \textit{sub-networks}.
By leveraging weight sharing, two-stage NAS dramatically accelerates the search process~\citep{bender-icml18,cai-iclr20} compared to classical NAS methods.

\textbf{Super-network Training.}  
Training a neural network involves iteratively updating the weight vector according to:  
\[
\rvw_{t+1} = \rvw_t + \lambda \delta_t
\]
where $\lambda$ is the learning rate and $\delta_t = \nabla_{\rvw} \mathcal{L}_{train}$ represents the gradient of the training loss with respect to all weights.  $\mathcal{L}_{train}$ in the case of language models refers to the negative log-likelihood loss for next-token-prediction. 

When training a super-network, it is essential to ensure that sub-networks perform well independently~\citep{yu-eccv20}, preventing co-adaptation among them. This guarantees that sub-networks retain strong performance when extracted from the super-network.  

To mitigate co-adaptation, the \textit{sandwich rule}~\citep{yu-iclr19} modifies the update step:  
\begin{equation}\label{eq:sandwich}
   \delta_t = \nabla_{\rvw} \mathcal{L}_{train}(\rvw_\Theta) + \sum_{i=1}^k \nabla_{\rvw} \mathcal{L}_{train}(\rvw_{\vtheta_i})
\end{equation}
where $\vtheta_i \sim \Theta$ is sampled uniformly from the search space.  

The sandwich rule aggregates gradients from the full super-network $\rvw_\Theta$ and $k$ random sub-networks $\rvw_{\vtheta_i}$. Intuitively, it balances computational effort across larger and smaller models, ensuring that the sub-networks in a search space receive adequate coverage during training. This enhances the prunability of the super-network, allowing sub-networks to achieve competitive performance in isolation. Previous work~\citep{yu-eccv20} also included the smallest sub-network $\vtheta_{min}$ in the update step, however, we found it not contributing to the overall performance.

\textbf{In-place Knowledge Distillation.}  
The goal of in-place knowledge distillation is to ensure that sub-networks retain the predictive power of the super-network by forcing their predictions to closely match those of the super-network.  

Drawing inspiration from traditional knowledge distillation techniques~\citep{hinton-arxiv15}, in-place knowledge distillation~\citep{yu-iclr19} minimizes the Kullback-Leibler (KL) divergence between the predictions of the super-network, $\pi_{\Theta}$, and a sub-network, $\pi_{\vtheta}$. The loss function for a single sub-network is defined as:  
\begin{equation}\label{eq:kd_loss}
\mathcal{L}_{KD} = \mathcal{L}_{train} + \KL\left(\sigma_{SM}\left(\frac{\pi_{\Theta}}{T}\right), \sigma_{SM}\left(\frac{\pi_{\vtheta}}{T}\right)\right),
\end{equation}
where $\sigma_{SM}(\cdot)$ denotes the softmax function, and $T$ is the temperature parameter that controls the smoothness of the output distribution. Note that $\KL$ can be replaced by different loss functions such as the reverse-kl, cosine-similarity or mean-squared error~\citep{xu2024survey} (see also Appendix~\ref{app:kd_loss}).

Since the sandwich rule (Equation~\ref{eq:sandwich}) already involves forward passes through the full super-network, the predictions $\pi_{\Theta}$ are readily available during training. This allows us to compute $\mathcal{L}_{KD}$ in-place, without any additional computational overhead.

\section{Scalable Model Compression Using Neural Architecture Search}\label{sec:method}

We now present our data driven approach to automatically compress pre-trained LLM. 
First, we describe a search space to compress decoder-based transformers in Section~\ref{subsec:search_space}.
Section~\ref{subsec:sampling} outlines our sampling strategy that allocates updates to a calibrated set of high-performing sub-networks. 
In Section~\ref{subsec:importance}, we study importance-based sorting mechanisms to improve sub-network initialization before super-network training. 
Section~\ref{subsec:peft} discusses how we can integrate our method with parameter-efficient fine-tuning techniques, such as LoRA~\citep{hu2021lora}, to improve scalability and efficiency. 

\subsection{Search Spaces}\label{subsec:search_space}

Following \citet{sukthanker-arxiv24}, we factorize the search space \(\Theta\) as:  
\[
\Theta = \Theta_{d_{model}} \times \Theta_{H} \times \Theta_{d_{head}}  \times \Theta_{r} \times \Theta_{L}
\]
where \(\Theta_{d_{model}}\) defines the range for the embedding dimension, \(\Theta_{H}\) specifies the number of attention heads, \(\Theta_{d_{head}}\) represents the head size, \(\Theta_{r}\) controls the expansion ratio for the feed-forward network (FFN) hidden dimension such that $r = \nicefrac{U}{d_{model}}$, and \(\Theta_{L}\) corresponds to the total number of transformer blocks.  
Compared to other pruning methods~\citep{ma2023llm}, which for example only prune single heads or have different number of heads per layer, we are more flexible in the design of our search space.
For example, to achieve actual latency gains (see Section~\ref{sec:experiments}) we change parameters in powers of two and keep the number of heads fix across layers.

Table~\ref{tab:searchspaces-all} in the Appendix shows the ranges for the different parameters for our search space  (dubbed \textit{Joint-Space}). 
We also consider two additional search spaces that either fix the head size (dubbed \textit{Fixed-Head-Size-Space}) or fix the total number of heads (dubbed \textit{Fixed-Head-Space}), which has been considered by previous approaches, such as Minitron~\citep{muralidharan2024compact}.
We provide a more detailed comparison in Section~\ref{subsec:importance}.

\begin{figure}
    \centering
    \includegraphics[width=\linewidth]{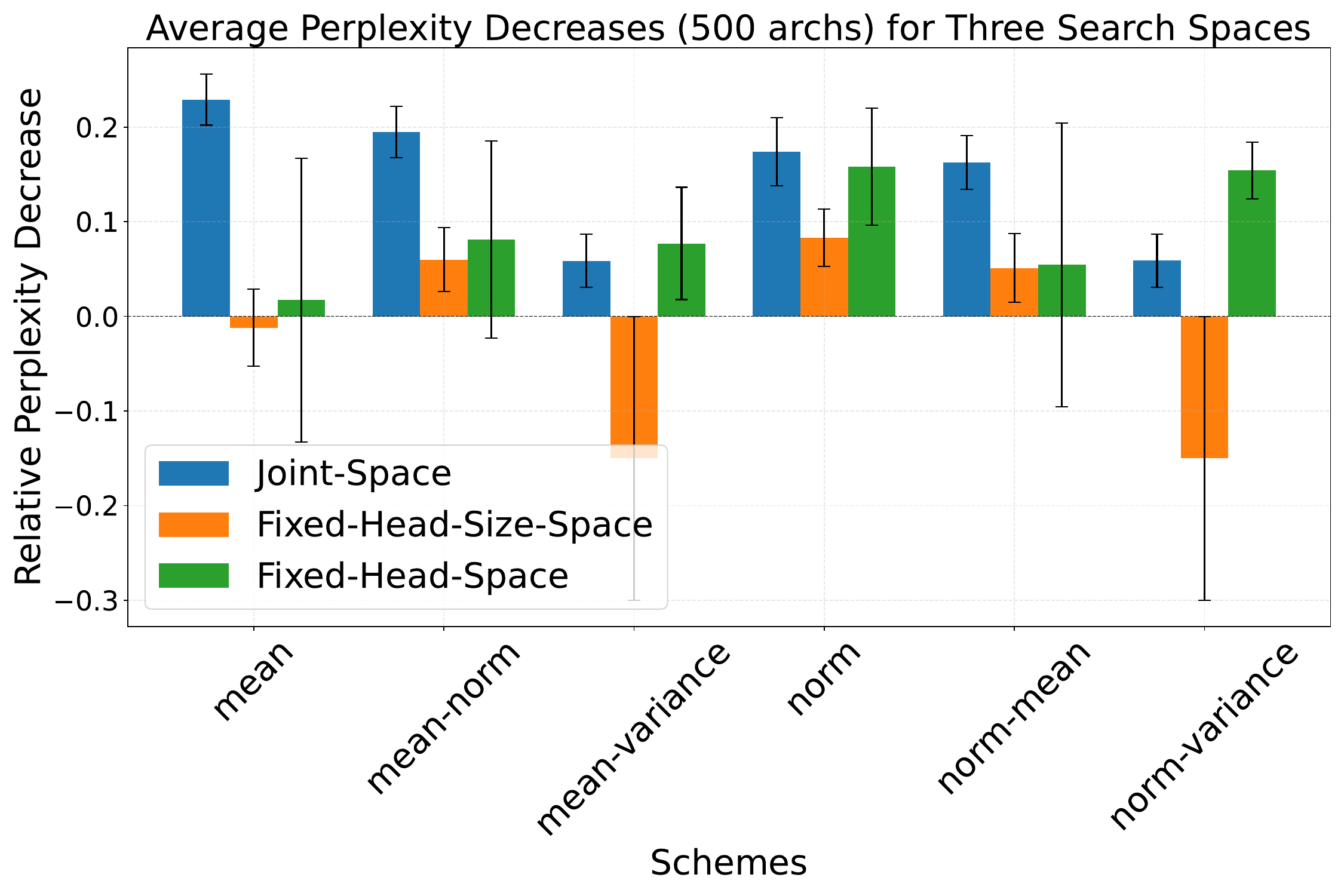}
    \caption{Block importance scheme}
    \label{fig:imp_sorting_bi}
\end{figure}



%
%

\subsection{Sampling Distribution}\label{subsec:sampling}

\begin{figure}
  \centering
  \begin{subfigure}{0.49\linewidth}
      \includegraphics[width=\linewidth]{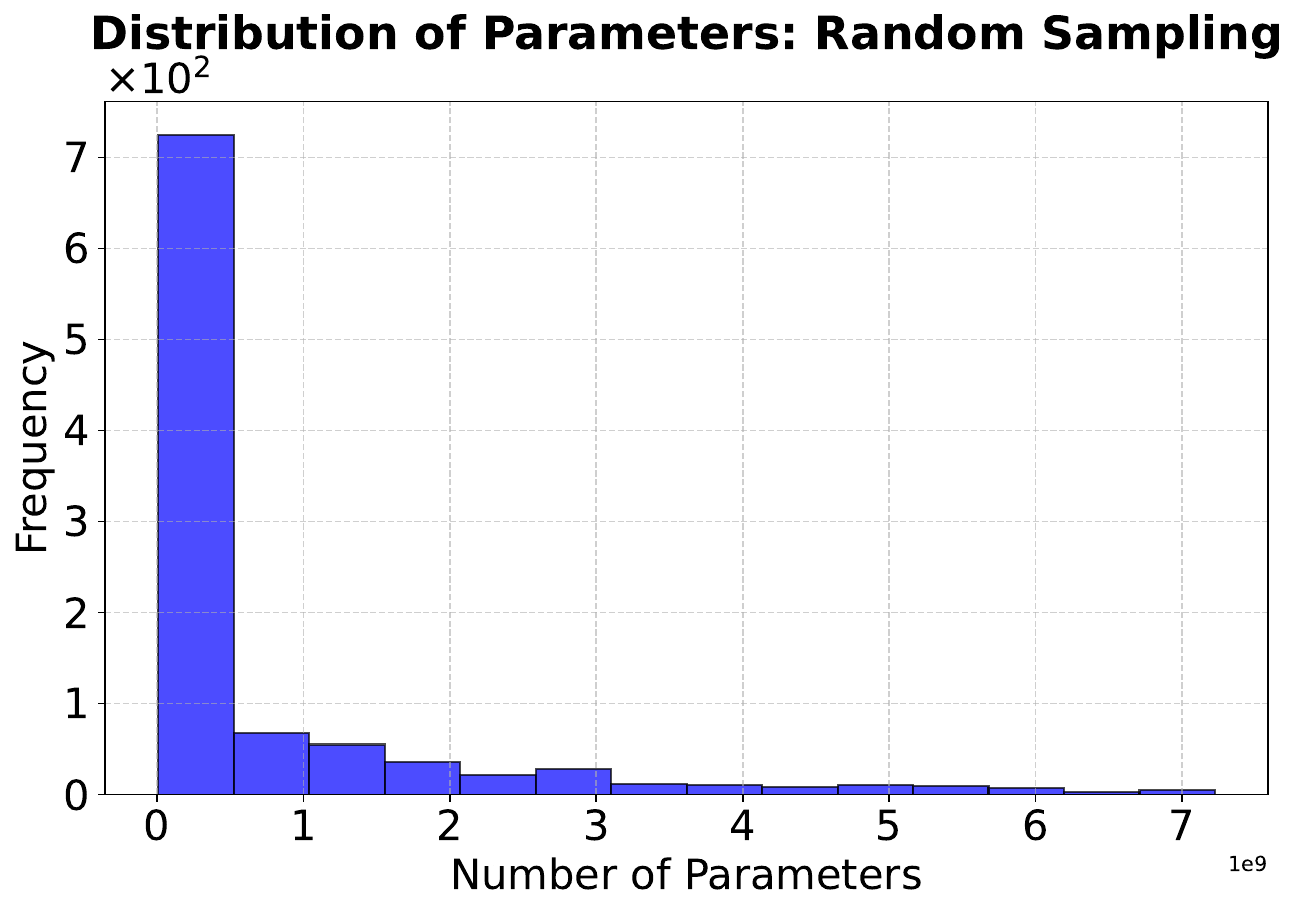}
      \caption{}
      \label{fig:random_sampling}
  \end{subfigure}
  \hfill
  \begin{subfigure}{0.49\linewidth}
      \includegraphics[width=\linewidth]{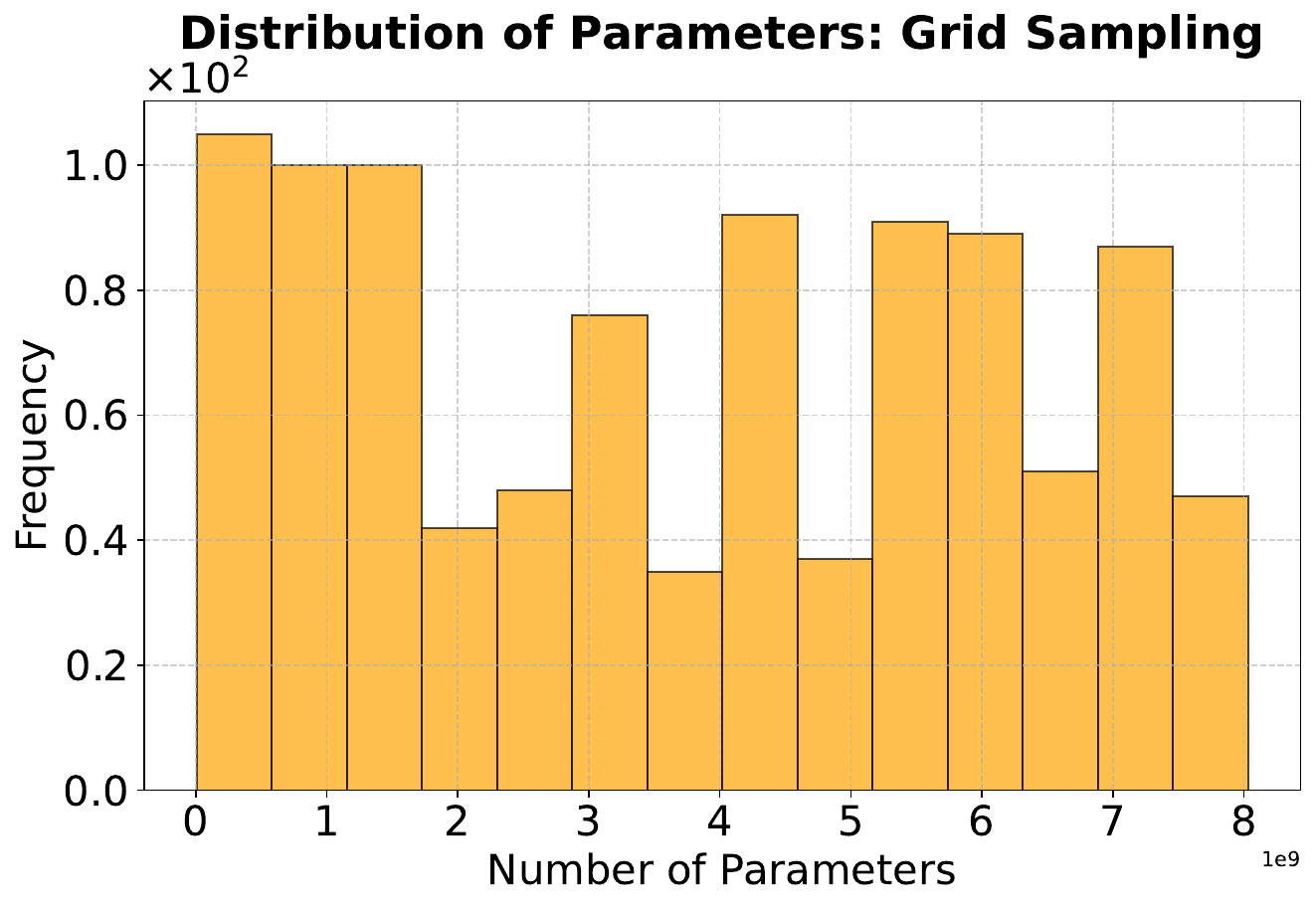}
      \caption{}
      \label{fig:grid_sampling}
  \end{subfigure}
  
  \caption{Parameter count distribution for sub-networks derived from the \textit{Joint-Space}. \textbf{(a)} sampled using \textbf{random-sampling} scheme. \textbf{(b)} sampled according to \textbf{grid-sampling} scheme. We can see that sampling randomly tends to over-sample tiny models that are not capable of achieving reasonable performance.}
  \label{fig:sampling_methods}
  \vspace{-4mm}
\end{figure}

The sandwich rule (see Section~\ref{subsec:background_nas}), samples sub-networks uniformly at random, from the search space during each update step.
However, this results in a skewed distribution that favors smaller sub-networks.
For example, Figure~\ref{fig:random_sampling} shows the distribution of parameter counts for 1000 sub-networks of a Llama-3.1-8B model sampled uniformly at random from our search space. 
As illustrated about 700 out of 1000 sub-networks have fewer than 500M parameters. 
These tiny models typically lack the capacity to learn effectively, yet they receive a disproportionate number of updates during super-network training. 
This skew towards smaller models only worsens as the overall model size increases, leading to inefficient use of computational resources.
Reducing the search space $\Theta$ to mitigate this imbalance is challenging, as it introduces bias and compromises the exploration-exploitation trade-off.

To ensures that sub-networks across a broader parameter range receive adequate training, we propose a different sampling strategy that overcomes the imbalance caused by uniform random sampling and enhances overall model performance. 
We allocate more update steps to a curated set of sub-networks, $\sH$, with parameter counts that are uniformly distributed across the search space:  
\[
   \delta_t = \nabla_{\rvw} \mathcal{L}_{train}(\rvw_\Theta) + \sum_{\vtheta_i \sim \sH} \nabla_{\rvw} \mathcal{L}_{train}(\rvw_{\vtheta_i}) \tag{1} \label{eq:sandwich_update}
\]
Let $\vtheta_{min}$ and $\vtheta_{max}$ denote the smallest and largest sub-networks in $\Theta$, with parameter counts \(params_{min} = params(\vtheta_{min})\) and \(params_{max} = params(\vtheta_{max})\), respectively. We discretize this parameter range into a grid $\sG = \{params_{min}, \dots, params_{max}\}$ with $K$ equally spaced bins.  
Now, to construct the set $\sH$, we apply rejection sampling to select sub-networks that fall within each bin. For each bin \(g_i \in \sG\), we sample $M$ sub-networks:  
\[
\sS_i = \{\vtheta_1, \dots, \vtheta_M \} \quad \text{such that} \quad g_{i-1} \leq params(\vtheta_i) \leq g_i
\]
We then compute the weight magnitude of each sub-network $mag(\vtheta_i) = \| \rvw_{\vtheta_i} \|_1
$.
From the set $\sS_i$, we select the sub-network with the highest weight magnitude:  
$
\hat{\vtheta}_i = \argmax_{\vtheta \in \sS_i} mag(\vtheta)
$
resulting in a curated grid of sub-networks
$
\sH = \{\hat{\vtheta}_0, \dots, \hat{\vtheta}_K\}
$.
Now, in each update step, we sample sub-networks $\vtheta_i \in \sH$ as described in Equation~\ref{eq:sandwich}.

We show in the next Section~\ref{subsec:importance}, how we can further calibrate the architectures based on sub-network magnitude after sorting the super-network.
This ensures that, our sampling procedure favors more effective architectures following an intuition similar to \citet{han-neurips2015}
\subsection{Importance sorting}\label{subsec:importance}


Given a configuration \(\vtheta = [768, 8, 32, 2, 9]\), the sub-network is constructed by selecting the first entries for each component, for example the first \(d_{model} = 768\) entries from the embedding vector. 
This convention ensures a bijective mapping between the search space \(\Theta\) and its sub-networks, as outlined by \citet{klein-tmlr24}.
However, it also might lead to less effective sub-networks.  

\begin{figure}
    \centering
    \includegraphics[width=0.45\textwidth]{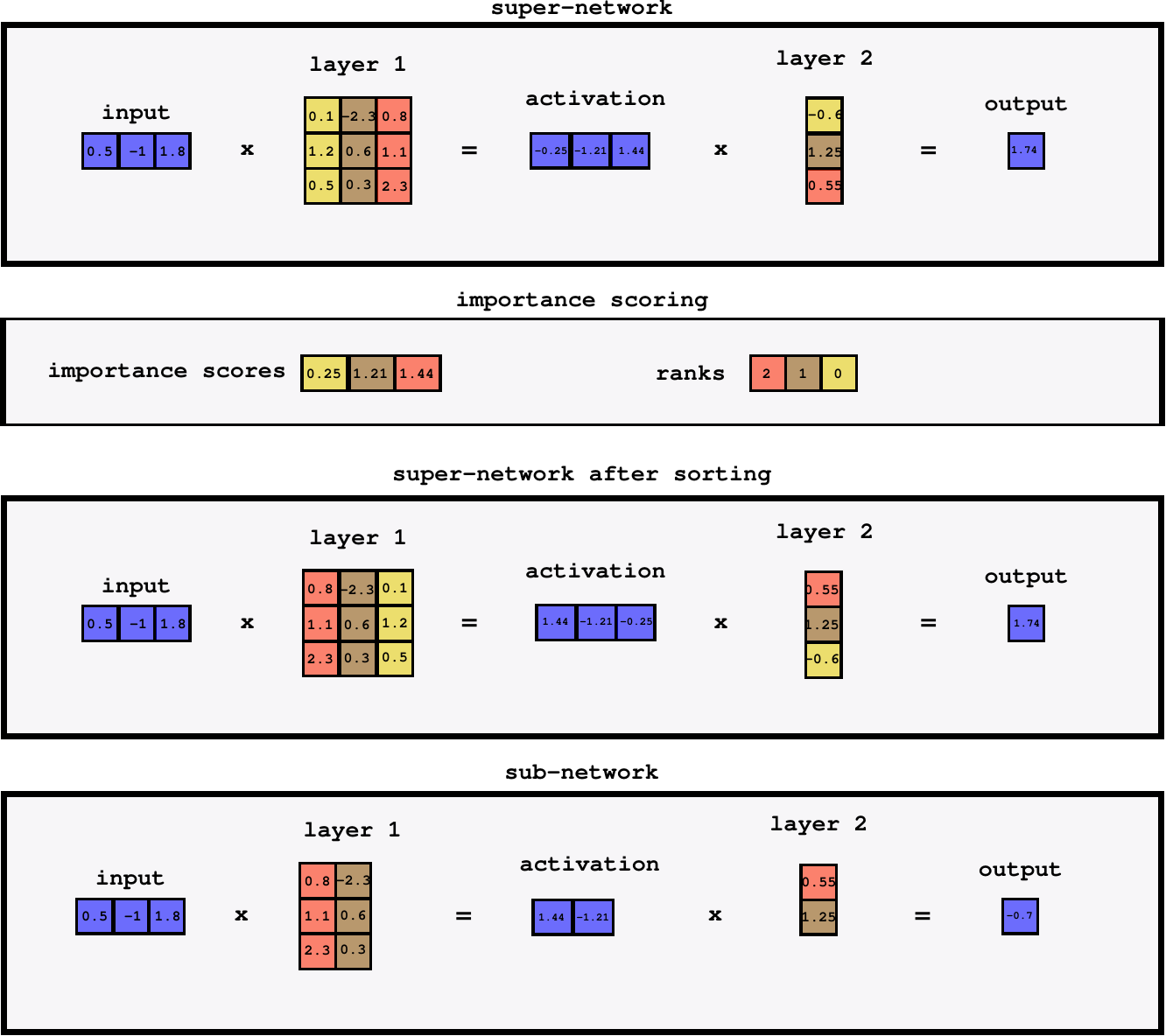}
    \caption{Illustration of importance sorting for a simple 1-layer FFN with 3 units. Reshuffling the hidden units does not change the final output.
    After sorting, we extract a sub-network with one 2 hidden units.}
    \label{fig:sorting_illustration}
\end{figure}

To mitigate this, we can sort the different components of a transformer based on their importance.
Figure~\ref{fig:sorting_illustration} shows an illustration for a simple FFN. 
This ensures that the first elements selected during sub-network extraction correspond to the \textit{most important} elements, as ranked by the chosen importance score.
We adopt the dimension-wise importance scoring and sorting proposed by~\citet{muralidharan2024compact}, which uses the activation of a component as proxy for its importance, with one noticeable difference.
Compared to~\citet{muralidharan2024compact}, for LLMs using Grouped Query Attention (GQA), such as Llama 3.1, we introduce a key modification and compute importance at the group-level instead of the head level to reflect the structural dependencies. Importance sorting in conjunction with the calibrated grid leads to significant improvements in sub-network quality as seen in Figure~\ref{fig:sampling_schemes} in the appendix.



Given a batch as input $\mX \in \mathbb{R}^{B\times T \times d_{model}}$ after applying the embedding layer $\mW^{emb}$ we compute the following scores for each component, where $B$ corresponds to the batch dimension and $T$ corresponds to the sequence length dimension, and $abs$ corresponds to the absolute value function:
\begin{itemize}
\vspace{-4mm}
\item For a neuron $i \in \{1, ..., U\}$ in a FFN layer $l$, we compute its importance by: 
        \( F^{(i)}_{FFN_l} = \nicefrac{1}{B}\sum_B \left( \nicefrac{1}{T}\sum_T \mathbf{X}\mW_1^l[:, i]  \right)\) where $\mathbf{W}_1^l[:, i]$ corresponds to all weights of neuron $i$ in layer $l$.
\item Similarly for each neuron $i \in \{1, ..., d_{model}\}$ in the embedding layer we compute 
        \( F^{(i)}_{\text{emb}} =  \nicefrac{1}{B}\sum_B  \left( \nicefrac{1}{T}\sum_T \left( \text{RMSNorm}(\mathbf{X}[:, :, i]) \right) \right)\). Specifically we perform mean absolute aggregation over output of every RMSNorm layer as defined in \ref{sec:background}.
\item For GQA layers we compute the importance per group $g \in \{1, ..., G\}$ of heads as :\[
F^{(g)}_{\text{GQA}} = \nicefrac{1}{B}\sum_B \Big(
   \nicefrac{1}{T}\sum_T  \Big\|
    \text{Attn}\big(\mQ_g, \mK_g, \mV_g \big) \Big\|_2 \Big)\]
\item For a block $l \in \{1, ..., L \}$ consisting of a GQA and a FFN layer with RMS layer normalization in between, we compute the score:
        \(
        \text{F}_{block}^{(l)} = 1 - \nicefrac{1}{B}\sum_B \Big(
   \nicefrac{1}{T}\sum_T  \left( \frac{\mX_{l}^T \mX_{l+1}}{\|\mX_{l}\|_2 \|\mX_{l+1}\|_2} \right)\Big) \;
\) where $\mX_l$ is the input to block $l$ and  $\mX_{l+1}$ the output.
\end{itemize}

Figure~\ref{fig:imp_sorting_bi} compares different schemes to aggregate importance across batch and sequence length (see Appendix~\ref{subsec:imp_sorting_extended} for more Details). Mean aggregation achieves the greatest perplexity drop, while \textit{joint-space} benefits most from importance sorting. We use these two choices for the rest of the paper.

\subsection{NAS with Parameter Efficient Fine-Tuning}\label{subsec:peft}

Full Fine-Tuning of LLMs is often prohibitively expensive, both in terms of computational cost and GPU memory consumption. 
To enable the fine-tuning of larger models, we adopt parameter-efficient fine-tuning  techniques. In particular, we build upon LoNAS~\citep{munoz2024lonas}, introducing two key modifications. Firstly, we apply LoRA to the embedding layer and the $\mW^{Q},\mW^{K},\mW^{V}$ matrices. This contrasts with LoNAS, which restricts LoRA to attention and FFN layers. Second, since our method also selects the total number of transformer blocks, we dynamically prune obsolete LoRA modules as layers are dropped. This extension allows us to reduce the embedding dimension and eliminate entire transformer blocks, leading to further latency improvements compared to LoNAS. We refer to the joint tuning of multiple architectures using weight sharing, importance sorting and calibrated architecture selection as  \textit{ours-weight-sharing}. We also go a step further and fine-tune each of the architectures in the calibrated grid independently with LoRA. We refer to this as \textit{ours-no-weight-sharing}, as it differs primarily in allowing each sub-network to adapt independently, avoiding the co-adaptation effects introduced by two-stage NAS (see Section~\ref{subsec:background_nas}), while introducing about $\times5$ more compute.  
\ifx\skiptodos\undefined
\fi

\section{Experiments}\label{sec:experiments}

We now present the empirical evaluation of our proposed method. 
Section ~\ref{subsec:exp_datasets} describes the datasets we use for importance sorting, fine-tuning and evaluation.
We describe the details of our fine-tuning pipeline in Section~\ref{subsec:exp_details}.
Section~\ref{subsec:exp_pruning} shows the comparison of the proposed approach against state-of-the-art pruning methods from the literature
as well as a simple, but effective, baseline which fine-tunes smaller models with weights extracted from the super-network. 
We also perform a more thorough ablation for different in-place KD losses and sampling schemes in Appendix~\ref{sec:ablations}.

\subsection{Datasets}\label{subsec:exp_datasets}

\textbf{Calibration Dataset for Importance Sorting}
To compute the importance scores as described in Section~\ref{subsec:importance}, we curate a diverse calibration dataset by sampling 5,000 text sequences (each of length 512) from each dataset, ensuring broad task and domain coverage. Check Appendix~\ref{subsec:imp_sorting_extended} for details on the dataset.
Neuron, layer, and head importance are computed using a subset of 400 samples drawn from this curated dataset, making the importance computation cheap and efficient.

\textbf{Fine-tuning Datasets}
We fine-tune LLaMA-3.1-8B for \textit{commonsense reasoning} tasks, by replicating the experimental setup of \citet{ma2023llm}\footnote{https://github.com/horseee/LLM-Pruner} using an instruction-tuning dataset of 2.58M instructions released by~\citet{lamini-lm}.
For \textit{arithmetic tasks} we adopt the setup from \cite{hu2023llm} and fine-tune LLaMA-3.1-8B using a combined dataset of seven arithmetic reasoning tasks with LM-generated chain-of-thought steps (\textbf{MATH10K}). We use the same calibrated grid of sub-networks for our sampling process, however, we perform two separate super-network fine-tuning runs, one for the commonsense reasoning domain and one for the math domain. 

\begin{table*}[hbt!]
    \centering
   \resizebox{\textwidth}{!}{ \begin{tabular}{llcccccccccccccc}
       \toprule
        \textbf{Model-Size} & \textbf{Method} & \textbf{Params} & \textbf{Latency} & \textbf{ARC-C} & \textbf{ARC-E} & \textbf{MMLU} & \textbf{PIQA} & \textbf{HellaSwag} & \textbf{Winogrande} & \textbf{BoolQ}  & \textbf{Lambada} & \textbf{MathQA} & \textbf{ASDIV}& \textbf{GSM8K}& \textbf{Avg.}\\
        \midrule
        \multirow{6}{*}{4B-5B} 
            & LLMPruner       & 4.14B & 72.18 & 22.87 & 33.21 & 22.92 & \underline{59.85} & 30.11 & 51.62 & 41.65  & 2.17 &20.435 &  0.00& 0.00& 25.89\\
            & LaCo            & 4.32B & \uline{61.08}  & 24.49 & 35.816 & 24.45 & 54.62 & 31.08 & 51.06 &  41.16 & 0.19 &  21.80 &  0.0&  0.0 &25.88\\
            & SliceGPT        & 4.41B & 75.21 & 22.53 & 34.30 & 24.27 & 55.33 & 29.86 & 51.93 & 37.83 &  9.2 & 20.54 & 0.0 & \underline{1.44} & 26.11\\
            & random-init                   & 4.52B & 64.46 & 23.98 & 30.22 & 22.99 & 55.33& 25.14 & 49.01 & 38.07 & 0.47 & 21.11 & 0.0 & 0.0 & 24.21 \\
            & pre-trained-init          & 4.52B & 64.46 & 21.50 & 30.09 & 24.55 & 57.83 & 25.64 & 50.75 & 49.51 & 0.91 & 21.71 & 0.00 & 0.00 & 25.68 \\
            & \textbf{ours-weight-sharing} & 4.32B & \textbf{60.27} & \underline{25.77} & \underline{37.67} & \underline{24.56} & 58.76 & \underline{35.51} & \underline{52.64} & \underline{53.85} &  \underline{16.88} & \underline{23.82} &  \textbf{1.91}&  1.36 &  \underline{30.25}\\
            & \textbf{ours-no-weight-sharing} & 4.32B & \textbf{60.27} & \textbf{29.77}&\textbf{48.73} & \textbf{26.50}& \textbf{65.51} & \textbf{41.64} & \textbf{53.12} & \textbf{56.73} & \textbf{35.16} &\textbf{25.19}& 0.00 & \textbf{2.20} & \textbf{34.96} \\
        \midrule
        \multirow{6}{*}{5B-6B} 
            & LLMPruner       & 5.20B & 78.12 & 24.40 & 41.96 & 25.17 & 25.17 & 25.17 & 52.72 & 54.74  & 17.08 & 21.61 &  0.00& 0.00 & 26.18\\
            & LaCo            & 5.41B & \uline{77.23} & 28.41 & 42.92 & 24.63& 59.63 & 39.59 & \textbf{60.22} & 61.80 &  2.425 & 19.50&  \uline{0.34}&  0.0&30.86\\
            & SliceGPT        & 5.21B & 83.60 &  26.62& 39.27 & 25.15 & 56.80 & 33.49 & 53.83 &  37.83 & 17.74  & 21.11 & 0.00 & 1.75 &  28.51 \\
            & random-init                  & 5.25B & 87.16 & 24.31 & 30.51 & \underline{25.72} & 54.78 & 25.23 & 50.51 & 50.305 & 1.04 & 21.07 & 0.00 & 0.00 & 25.77 \\
            & pre-trained-init         & 5.25B & 87.16 & 24.32 & 28.32 & 25.70 & 54.35 & 25.74 & 51.30 & 39.08 & 1.55 & 23.69 & 0.00 & 0.00 & 24.91 \\

            & \textbf{ours-weight-sharing} & 5.41B & \textbf{76.29} & \uline{29.95} & \uline{46.42} & 24.08 & \uline{67.02} & \uline{46.36} & 54.14& \uline{62.26} &  \uline{32.10} & \textbf{26.16}&  0.0&  \underline{2.12}&  \uline{35.51}\\
            & \textbf{ours-no-weight-sharing}  & 5.41B & \textbf{76.29} & \textbf{33.96} & \textbf{56.73} & \textbf{31.82} & \textbf{72.80} &\textbf{53.39} & \uline{56.27} & \textbf{63.58} & \textbf{48.79} & \uline{25.70}  & \textbf{0.43} & \textbf{4.32}  & \textbf{40.71} \\
        \midrule
        \multirow{6}{*}{6B-7B} 
            & LLMPruner       & 6.15B & 108.24 & 32.76 & 55.39 & 26.51 & \uline{73.29} & 53.16 & 58.41 & 53.73 &  39.80 & 23.35 & \uline{0.48} &  0.001 &  37.90\\
            & LaCo            & 6.06B & 86.62 & 27.22 & 42.676 & 26.73 & 61.70 & 32.34 &59.59 & \uline{62.14} & 12.61& 25.93 &  \textbf{1.56} &  0.0 &   32.05\\
            & SliceGPT        & 6.02B & 90.01& 32.60 & 46.09 & 25.28 & 60.77 &43.45 & \uline{62.35} & 37.86 &  32.87 & 23.30 & 0.00 &  \uline{3.11}&  33.42\\
            & random-init                   & 6.45B & \textbf{82.55} & 22.78 & 31.02 & 25.00 & 56.20 & 25.24 & 50.67 & 37.52 & 0.68  & 21.67 & 0.00 & 0.00 & 24.62 \\
            & pre-trained-init           & 6.45B & \textbf{82.55} & 23.29 & 31.94 & 25.67 & 60.06 & 27.14 & 51.14 & 39.11 & 7.18& 24.29 & 0.00 & 0.15 & 26.36\\
            &\textbf{ours-weight-sharing} & 6.06 & \uline{86.22} & \uline{35.75} & \uline{60.06} & \uline{30.33} & 72.74 & \uline{58.84} & 61.17 & \textbf{67.19}  & \uline{46.21} & \uline{27.30}&  0.0&  1.667 &  \uline{41.93}\\
            & \textbf{ours-no-weight-sharing}  & 6.06B & \uline{86.22} & \textbf{39.50} & \textbf{61.24} & \textbf{36.93} & \textbf{75.41} & \textbf{62.97} & \textbf{62.67}  &  61.44 & \textbf{49.76} & \textbf{27.77} & 0.39 & \textbf{11.30} & \textbf{44.49} \\
        \midrule
        \multirow{6}{*}{7B-8B} 
            & LLMPruner       & 7.20B & 192.71 & 45.82 & \uline{71.84} & 43.60 & \uline{78.29} & \uline{73.69} & 72.45 & 64.16 & 58.74 & 28.81 & \textbf{0.95} & 0.02&  48.94\\
            & LaCo        & 6.28B\footnote{Since LaCO drops entire layers based on a cosine similarity threshold, this is the largest possible parameter size after dropping 8 layers with cosine similarity of 0.4} & \textbf{89.89} & 27.39 & 39.85 & 43.98 & 60.55 & 31.17 &55.56& 37.52 &  5.45 & 26.43 &  \textbf{0.95}&  0.0&  29.90\\
            & SliceGPT        & 7.21B & 107.65 & \textbf{52.13} & 68.90 & 45.49 & 70.29 & 65.99 &\uline{73.16} & 70.29 &  57.66 & 27.14 &  0.0&  6.67 &  48.88\\
             & random-init                   & 7.16B   & 112.10 & 23.46 & 31.10 & 23.92 & 55.77 & 25.19 & 49.64 & 52.72 & 0.23 & 21.17 & 0.09 & 0.00 & 25.75 \\
            & pre-trained-init           & 7.16B & 112.10 & 25.51 & 26.14 & 22.95 & 53.26 & 26.16 & 51.30 & 37.83 & 0.00 & 20.07 & 0.00 & 0.0 & 23.93 \\
            & \textbf{ours-weight-sharing} & 7.22B & \uline{103.85} & \uline{50.34}  & \textbf{73.90} & \uline{49.27} & 77.31 & 72.87 & 72.22 & \textbf{82.91}  & \uline{65.55} & \uline{34.00} & 0.13&  \underline{14.85}&  \uline{53.94}\\

            & \textbf{ours-no-weight-sharing}  & 7.22B & \uline{103.85} & 47.10& 66.16 & \textbf{53.90} & \textbf{78.45} & \textbf{74.37} & \textbf{73.40} & \uline{71.47} & \textbf{68.07} & \textbf{34.17}& \uline{0.61} & \textbf{43.14} & \textbf{55.53} \\
        \midrule
        \multirow{3}{*}{Semi-Structured} 
                    & Wanda 2:4       & 4.54B &  108.16 & 26.79 & 52.31 & 28.82 & 68.17 & 44.86 & 57.70 & 67.40 &  48.18 & 24.86 & 0.004 & 0.02&  38.10\\
            & SparseGPT 2:4   & 4.54B &  108.22& 34.13 & 57.28 & 36.58 & 69.42 & 53.94 & 61.96 & 73.03 &  55.09 &  25.83&  0.004 & 0.03 &  42.48\\
            & Magnitude 2:4   & 4.02B & 107.15 & 23.55 & 38.59 & 26.67 & 61.75 & 29.26 & 53.43 & 37.86 &  0.08 & 24.52 &  0.0 &  0.0&  26.88\\
        \bottomrule
    \end{tabular}}
    \caption{Comparison against different structured, semi-structured pruning methods and smaller fine-tuned models on commonsense-reasoning and math tasks. Note for \textbf{ours-weight-sharing}, we perform 2 separate supernet finetuning runs for commonsense and math reasoning tasks.}
    \vspace{-4mm}
    \label{tab:pruning-evaluation}
\end{table*}
\textbf{Evaluation Datasets} For commonsense reasoning, we evaluate our method on eight diverse reasoning datasets: BoolQ~\citep{clark2019boolq}, PIQA~\citep{bisk2020piqa}, HellaSwag~\citep{zellers2019hellaswag}, WinoGrande~\citep{sakaguchi2021winogrande}, ARC-e, ARC-c~\citep{clark2018think}, TruthfulQA~\citep{lin2022truthfulqa}, and MMLU~\citep{hendrycks2021measuring}.
These datasets encompass a range of tasks, including yes/no question answering (BoolQ), physical commonsense reasoning (PIQA), story continuation (HellaSwag), pronoun resolution (WinoGrande), scientific question answering (ARC-e/c), factual consistency assessment (TruthfulQA), and multi-domain understanding (MMLU).
To evaluate long-context generalization, we benchmark on LAMBADA.

Arithmetic reasoning is assessed using MathQA~\citep{amini2019mathqa}, GSM8K~\citep{cobbe2021gsm8k}, and ASDIV~\citep{miao2021diverse}.
We adopt the prompt template from \citet{hu2023llm}, applying string normalization by trimming whitespace.
Following standard evaluation protocols, we perform 5-shot evaluation for WinoGrande, 10-shot for HellaSwag, 25-shot for ARC-c, 5-shot evaluation for GSM8k, 5-shot evaluation for MMLU, and 0-shot evaluation for ARC-e, PIQA, BoolQ, LAMBADA, MathQA and ASDIV using LM-Eval-Harness~\citep{eval-harness}\footnote{https://github.com/EleutherAI/lm-evaluation-harness}.

\subsection{Experiment Details}\label{subsec:exp_details}
We adapt LoRA for super-network fine-tuning as described in Section~\ref{subsec:peft}.  
Specifically we set LoRA-rank to 32 and LoRA-alpha to 16 and LoRA-dropout to 0.05 during the fine-tuning procedure.
We place low-rank matrices on the $\mW^{Q},\mW^{K},\mW^{V}$ matrices of all the attention blocks and the embedding matrix $\mW^{emb}$, which we found to be useful in practice.
Furthermore, we fine-tune our super-network for a total for 3 epochs for both commonsense and math tasks with the learning rate annealed from 0.0002 to 6.0e-05 using cosine annealing.
We use Adam for fine-tuning with $\beta_1$ and $\beta_2$ set to 0.9 and 0.95, respectively.
We set the grid-size of our sampling schemes (see Section \ref{subsec:sampling}) to $K=22$. 

\subsection{Comparison to Pruning Baselines ans Smaller Models}\label{subsec:exp_pruning}
We compare against the following structural pruning methods from the literature:
\textbf{LLM Pruner}~\citep{ma2023llm}  uses gradient information to prune non-critical structural components, thereby preserving multi-task capabilities.
\textbf{LaCO}~\citep{yang2024laco} merges later layers into earlier ones, effectively decreasing model depth without altering the overall architecture.
\textbf{SliceGPT}~\citep{ashkboos-iclr24} computes data-dependent orthogonal projection matrices to rotate weight matrices which preserves the output of the model while reducing the overall size of the weight matrices. In addition we compare to different semi-structured 2:4 sparsity methods such as wanda~\citep{sun-iclr2024}, sparsegpt~\citep{frantar2023sparsegpt} and magnitude-based~\citep{han2015deep} pruning. 

As an additional ablation, we also compare our NAS approach against two simple baselines: \textbf{Random-init} samples a set of architectures using the sampling process described in Section~\ref{subsec:sampling} but initialize them randomly, and fine-tunes them independently for the same time budget as our method.  
\textbf{Pre-trained-init} uses the same set of networks but initialize their weights inherited from the pre-trained network and fine-tune them for the same duration as our method. 

Compared to our NAS approach, these two baselines incur approximately $\times5$ higher compute costs for an architecture grid of size $K=20$.
This already includes the extra overhead of approximately $\times4$ of the sandwich rule (see Equation \ref{eq:sandwich_update}) compared to vanilla SGD.

Our method achieves consistently lower latency compared to  \textbf{LLM Pruner}, \textbf{LaCO} and \textbf{SliceGPT} (see Table~\ref{tab:pruning-evaluation}).
Out of $11$ tasks, our weight-sharing based approach achieves $8.85\%$, $9.85\%$, $6.59\%$,  and $6.58\%$ improvements in average accuracy over all tasks compared to all three methods, for target sizes $\sim$ 4.5B, 5.2B, 6B and 7.2B, respectively. Similarly the latency gains relative to the best performing pruning methods are $14.93\%$, $0.94\%$, $22.02\%$ and $88.86\%$ across the four parameter sizes, respectively.
As expected, semi-structured baselines do not lead to inference speedups.  We profile all pruned models on a single A100 GPU with 8 CPU cores. 

Also, \textbf{random-init} and \textbf{pre-train-init} perform worse than our approach, despite requiring $\times5$ more compute.
This further emphasizes the importance of the different parts of our approach, i.e importance sorting and selection of sub-networks based on weight magnitude.

\section{Conclusions}

We present a NAS-based approach for compressing LLMs.
Unlike typical structural pruning methods, we search for a \textit{Pareto-optimal set} of sub-networks that balance performance and efficiency, measured in terms of parameter count or latency.
Across a range of common-sense reasoning and math tasks, we demonstrate that models compressed using our approach outperform state-of-the-art structural pruning methods at various target sizes.

Future work could focus on further reducing the computational overhead of NAS, ideally bringing it closer to that of standard SGD.
Additionally, we plan to explore combining our approach with recent advancements in synthetic data distillation to generate training data for our super-network fine-tuning process.

\section{Impact Statement}

Large language models are powerful but often suffer from excessive computational and memory demands, limiting their deployment in real-world applications. In this paper, we propose a novel automated pruning method that systematically identifies and removes redundant parameters while preserving model performance. Unlike traditional pruning approaches that rely on manual heuristics or fixed sparsity constraints, our method dynamically adapts to different model architectures and search space constraints, enabling more effective and generalizable compression.

Our approach significantly improves efficiency by reducing memory footprint and inference latency without requiring extensive retraining. This advancement might eventually make large-scale language models more accessible for deployment in resource-constrained environments such as edge devices and mobile applications. By automating the pruning process, our method simplifies model optimization, contributing to the broader goal of sustainable and scalable AI.


\bibliography{lib}
\bibliographystyle{icml2025}

\newpage
\appendix
\onecolumn

\section{Extended Related Work}\label{sec:related_app}
\textbf{Model compression} reduces the size or computational complexity of a neural network model while minimizing loss in performance.
It involves techniques such as quantization (reducing the precision of weights and activations), pruning (removing neurons or connections from a model), knowledge distillation (transferring knowledge from a large to a small network e.g. in the form of representations or outputs), low-rank factorization, efficient model design and NAS (see e.g. \citet{zhu-arxiv2023} for an overview).
In the following, we focus on pruning and NAS.

\textbf{Pruning} removes weights and connections of a network to reduce the number of parameters and accelerate inference.
Unstructured pruning removes arbitrary weights while structural pruning considers entire groups of parameters such as attention heads \cite{michel-nips19} or layers \citep{sajjad-2023} for removal which is better suited for model acceleration on hardware optimized for dense computation \citep{mishra-2021}.
Pruning and in particular structured pruning approaches can result in a loss of accuracy and most pruning methods include a retraining phase to recover as much accuracy as possible.
Recent work focused on pruning LLMs to tackle the particular challenges that come with their large number of parameters, the high computational complexity, and the often limited availability of data for retraining.
Methods such as ShortGPT \citep{men-arxiv2024} and LaCo \citep{yang-arxiv2024} use importance scores to prune or merge layers of LLMs.
SparseGPT \citep{frantar-arxiv23} approximates the optimal weights in a pruning mask using a row-wise iterative update scheme to do unstructured and semi-structure pruning of generative pre-trained transformers.
Wanda \citep{sun-iclr2024} extends magnitude pruning \citep{han-neurips2015} by including the activation values on a small calibration set to do unstructured and N:M structured pruning.
Flextron \citep{caiflextron} propose a procedure that allows to extract models for different deployment scenarios by first making by combining an elastic model (cf. \citet{cai-iclr20}) with methods from mixture of experts (see e.g. \citet{fedus-arxiv2022} for a recent review).
The router networks can take static information such as a target latency into account but also input-adaptive routing.
Probably the closest work to our approach is Minitron \citep{muralidharan2024compact}, which uses activation-based importance scores to prune models and knowledge distillation from an uncompressed teacher for retraining.
However, they need to considerably reduce the number of architectures that are compared to reduce training time.
In our work, we consider much larger search spaces and leverage one-shot NAS for efficient training including knowledge distillation and incorporate importance scores during the architecture sampling procedure.
This allows us to combine everything into a single one-step training procedure and calculate the full Pareto front instead of single architectures.

\textbf{Neural Architecture Search} automates the design of deep neural networks in a data-driven manner (see e.g. \citet{elsken-jmlr2019, white-arxiv23} for an overview).
NAS has been extended to a multi-optimization problem taking also efficiency on a target hardware platform into account such as latency or energy consumption \citep{elsken-iclr19, cai-iclr20, wang-ACL2020} making it closely related to model compression.
To tackle the enormous computational cost of early NAS methods \citep{zoph-iclr17a, real-aaai19}, weight-sharing based NAS \citep{saxena-neurips2016, bender-icml18} trains a single super-network from which the weights can be inherited to all architectures in a search space for performance evaluation without further training.
A particularly prominent approach to use NAS for model compression is two-stage NAS, which has a dedicated super-network training and multi-objective search stage \citep{bender-icml18, guo-eccv2020, cai-iclr20}.
Most two-stage methods use the notation of elastic layers \citep{cai-iclr20} that can dynamically adjust its size (e.g. width) during training.
The training of the supermodel is typically done as proposed in \citet{yu-ICCV2019} using the sandwich rule, which aggregates the gradients of multiple sub-networks from the super-network, as well as in-place distillation, which uses the outputs of the largest network in a super-network as targets for smaller ones.
Furthermore, \citet{tang-iccv2023} and \citet{wang-cvpr2021} proposed different strategies how to sample models from supermodel to better cover the Pareto front.
A detailed study of NAS for structural pruning has been conducted in \citet{klein-tmlr24} that shows that NAS is a competitive technique to other pruning approaches and highlights in particular the increased flexibility and automation potential of NAS methods that allow to estimate the full Pareto front \citep{cai-iclr20, sukthanker-arxiv2024b} instead of having to set a single threshold for pruning.
However, even two-stage NAS methods still have a large computational overhead compared to regular model training.
To further reduce the computational complexity of NAS, several works proposes to leverage pre-trained weights as well as parameter efficient fine-tuning methods.
InstaTune \citep{sridhar-iccv2023} uses a pre-trained model to initialize and train a super-network on a fine-tuning task.
LoNAS \citep{munoz2024lonas} freezes the weights of a pre-trained backbone and introduces elastic LoRA adapters \citep{hu-arxiv21}.
Similarly, Shears \citep{munoz-arxiv2024} combines unstructured pruning of a pre-trained model with NAS to search for elastic LoRA adapters to mitigate the performance loss of pruning.

\paragraph{Knowledge distillation (KD)} is a widely used technique for compressing LLMs by transferring knowledge from a larger teacher model to a smaller student, aiming to preserve performance while reducing computational overhead~\citep{hinton-arxiv15,xu2024survey}. However, applying KD to LLMs is often computationally expensive, requiring either simultaneous teacher-student memory usage or precomputed logits. To mitigate this, \textit{in-place knowledge distillation}~\citep{yu2019universally,caionce,ruiz2019adaptative,guerra2020switchable} enables efficient KD by distilling knowledge from a larger network to its smaller sub-networks during training. In this work, we leverage in-place knowledge distillation in conjunction with parameter-efficient fine-tuning techniques, such as LoRA~\citep{hu2021lora}, to further reduce computational costs while maintaining strong downstream performance.
\section{Experimental details}
In this section we provide details on the importance sorting schemes, sampling schemes and the LoRA hyperparameters.
\subsection{Fine-tuning Pipeline} Our fine-tuning hyperparameters for Llama-3.1-8B largely follow \emph{litgpt}\footnote{\url{https://github.com/Lightning-AI/litgpt/blob/main/config_hub/finetune/llama-3.1-8b/lora.yaml}}, in addition to core differences described in Section~\ref{subsec:exp_details}. We fine-tune all models for 3 epochs on a single A100 GPU, for about 48 GPU hrs for commonsense reasoning tasks and for about 10 hrs for math tasks. 
\subsection{Details on Importance Sorting }
\label{subsec:imp_sorting_extended}
\begin{figure*}[t]
\centering
  \begin{subfigure}{0.47\textwidth}
      \includegraphics[width=\textwidth]{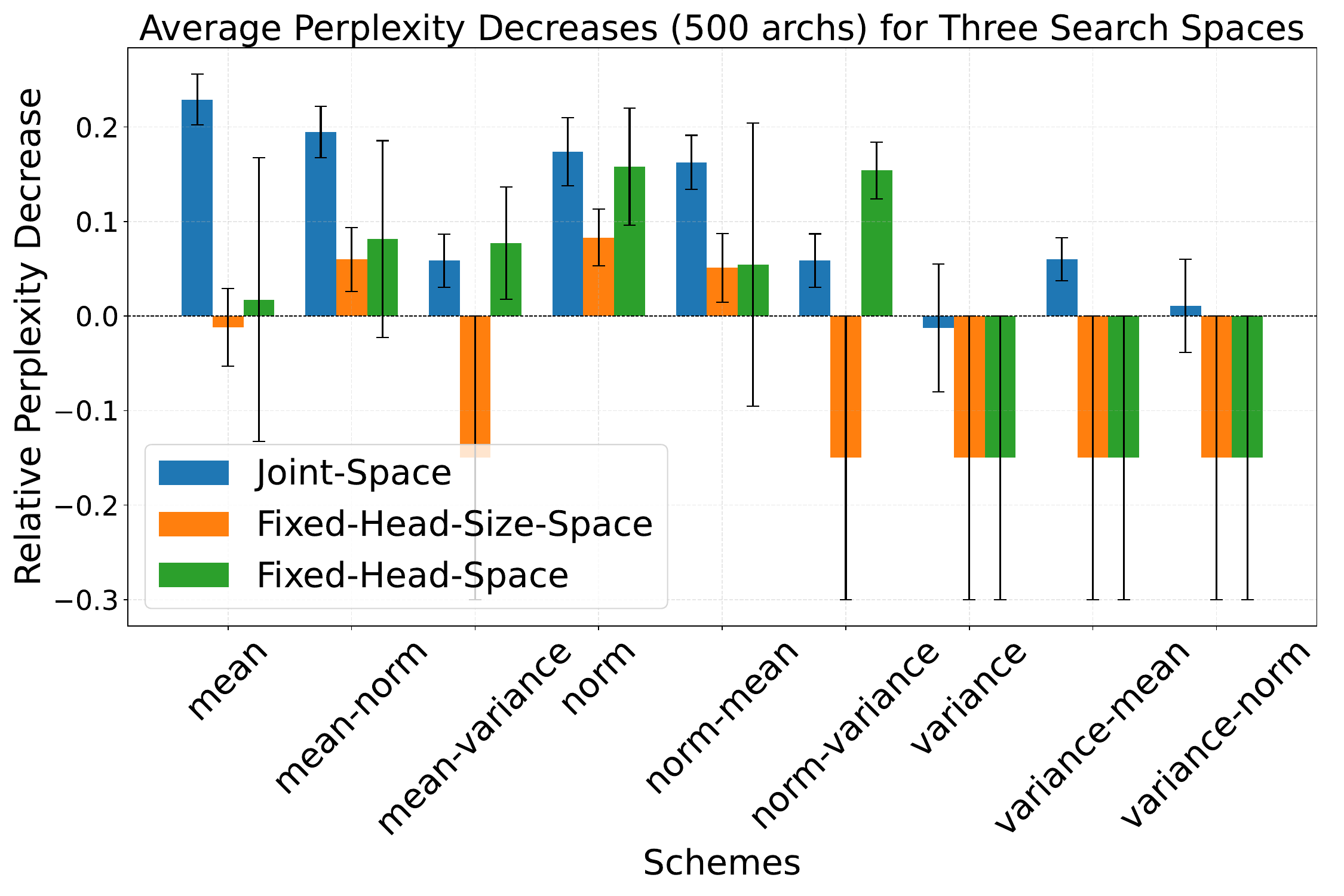}
      \caption{Block Importance}
  \end{subfigure}
  \hfill
  \begin{subfigure}{0.47\textwidth}
      \includegraphics[width=\textwidth]{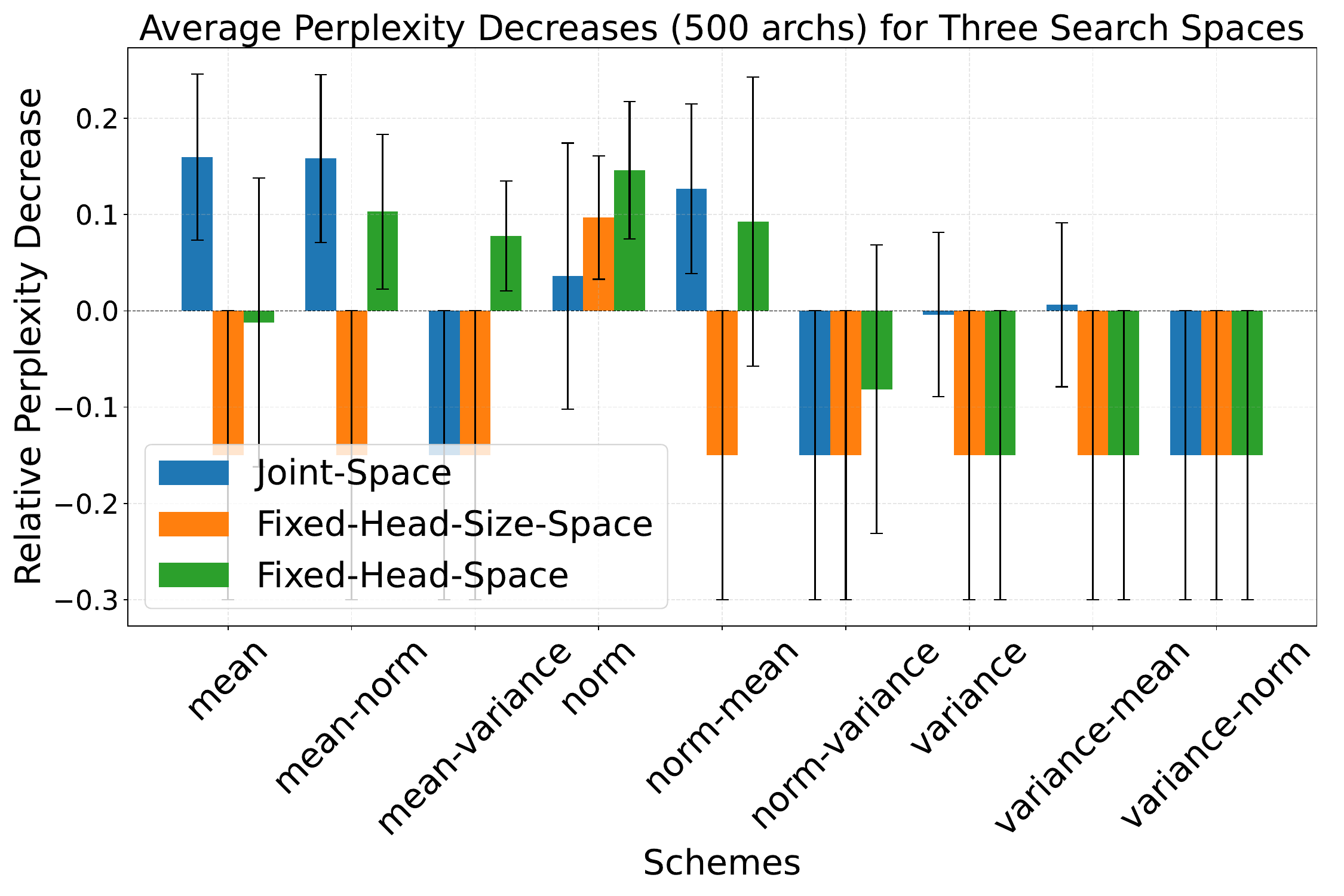}
      \caption{Layer Drop}
  \end{subfigure}

  \caption{Importance Sorting Relative Perplexity Decrease for different aggregation schemes}
  \label{fig:importance_sorting_all}
\end{figure*}

\paragraph{Calibration Dataset for Importance Computation} Our calibration dataset includes \textit{C4} \citep{raffel2020exploring}, \textit{Wikitext-103} \citep{merity2016pointer}, and \textit{OpenWebText} \citep{gokaslan2019openwebtext} to capture diverse internet text, \textit{Alpaca} \citep{alpaca2023} for instruction-following tasks, \textit{Commonsense170k} \citep{wang2019does} for reasoning tasks, \textit{TinyStories} \citep{eldan2023tinystories} for simplified narratives, and \textit{GSM8K} \citep{cobbe2021gsm8k} for math problem-solving.
Additionally, we incorporate \textit{Lambada} \citep{paperno2016lambada} for narrative completion, \textit{XSum} \citep{narayan2018don} and \textit{CNN/DailyMail} \citep{hermann2015teaching} for summarization, \textit{Samsum} \citep{gliwa2019samsum} for dialogue summarization, and \textit{Yelp Reviews} \citep{zhang2015character} for sentiment classification to further diversify the set. 
\paragraph{Different possible search space choices} We study importance sorting on three different search spaces described in Table~\ref{tab:searchspaces-all}. We observe that the search space consisting of jointly searching over number of heads and head size, tends to out-perform other search spaces with the head-size or the number of heads fixed (see Figure~\ref{fig:importance_sorting_all}). 
\paragraph{Different importance aggregation schemes} Following~\citet{muralidharan2024compact}, we investigates different methods \( \text{agg}_B \) and \( \text{agg}_S \),
to aggregate the importance of a component across the batch size and the sequence length, respectively. Unlike Minitron, which restricts analysis to a fixed architecture grid, our study extends to a larger set of 500 distinct architectures across three different search spaces. This broader scope aims to ensure a more comprehensive and robust comparison of aggregation strategies across diverse model configurations. For example \emph{norm-mean}, aggregation scheme refers to aggregating with norm across the batch dimension and aggregating with mean across the sequence length dimension. We observe that \emph{mean-mean} or simply \emph{mean} aggregation performs best for the \emph{joint-space}, followed by \emph{mean-norm} and \emph{norm-norm} schemes, while \emph{variance} based schemes perform poorly across search space types as seen in Figure~\ref{fig:importance_sorting_all}.
\paragraph{Block Importance v/s Block Drop Scheme} In addition to block-importance scheme defined in Section~\ref{subsec:importance}, we also study the \emph{block-drop} scheme as studied in~\citep{muralidharan2024compact}, which drops layers and gives a higher score to layers with largest impact on perplexity. However, we observe that block importance scheme yields higher average improvement over the joint search space. 
\paragraph{Evaluating Importance Sorting.}  
We evaluate three types of search spaces: one with a fixed number of attention heads, one with a fixed head size, and a joint search space, as shown in Table~\ref{tab:searchspaces-all}, which allows flexibility in both head count and head size.  
To provide a robust measure of the gains achieved by different importance sorting schemes, we introduce the concept of Relative Perplexity Decrease (RPD), defined as:  
\[
\text{RPD} = \frac{1}{N} \sum_{i=1}^{N} \frac{\left( \text{PPL}_{i}^{\text{before}} - \text{PPL}_{i}^{\text{after}} \right)}{\text{PPL}_{i}^{\text{before}}}
\]
where \(N\) represents the total number of sampled architectures.  

Figure~\ref{fig:importance_sorting_all} shows the results of applying these aggregation schemes across the three search spaces, evaluating their RPD over 500 randomly sampled architectures from the search spaces described in Table~\ref{tab:searchspaces-all}. Figure~\ref{fig:importance_sorting_all} shows that it is indeed useful to search in the joint space of number of heads $\mathcal{\theta}_{H}$ and $\mathcal{\theta}_{d_{head}}$, i.e. the \emph{Joint-Space}, benefits the most from importance sorting. Furthermore, we find that Block Importance (BI) works favorably compared to Block Drop (BD), amongst the layer importance choices presented in Section~\ref{subsec:importance}. We also find that the \emph{mean-mean} scheme to work the best, followed by \emph{norm-mean} and schemes involving variance do not work favourably as seen in Figure~\ref{fig:importance_sorting_all}. 

\begin{table}[h!]
    \centering
    \renewcommand{\arraystretch}{1.5}
    \resizebox{\textwidth}{!}{%
    \begin{tabular}{cccccc}
        \toprule
        \textbf{Search Space} & \textbf{\(\Theta_{d_{model}}\)} & \textbf{\(\Theta_{H}\)} & \textbf{\(\Theta_{d_{head}}\)} & \textbf{\(\Theta_{r}\)} & \textbf{\(\Theta_{L}\)} \\
        \midrule
        \textbf{Joint-Sapce} & $\{ 2^i \mid i \in [5, \log_2(d_{model})] \}$ & $\{8, 16, 32\}$ & $\{8, 16, 32, 64, 128\}$ & $\{1.0, 2.0, 3.0, 3.5\}$ & $\{1, ..., L\}$  \\
        \midrule
        \textbf{Fixed-Head-Size} & $\{ 2^i \mid i \in [5, \log_2(d_{model})] \}$ & $\{8, 16, 32\}$ & $\{128\}$ & $\{1.0, 2.0, 3.0, 3.5\}$ & $\{1, ..., L\}$  \\
        \midrule
        \textbf{Fixed-Head} & $\{ 2^i \mid i \in [5, \log_2(d_{model})] \}$ & $\{32\}$ & $\{8, 16, 32, 64, 128\}$ & $\{1.0, 2.0, 3.0, 3.5\}$ & $\{1, ..., L\}$ \\
        \bottomrule
    \end{tabular}}
    \caption{Specification for Joint, Fixed-Head, and Fixed-Head-Size Search Spaces for Llama 3.1-8B}
    \label{tab:searchspaces-all}
\end{table}
\subsection{Sampling} We use uniform size of $K=22$ for the architecture grid defined in \ref{subsec:sampling}. Furthermore when doing rejection sampling based on parameter count to obtain architectures in different parameter bins, we allow for at most 10000 architecture sampling trials.

\section{Ablations}
\label{sec:ablations}
\subsection{Impact of Sampling Schemes}
\begin{figure}[t]
    \centering
    \includegraphics[width=.49\linewidth]{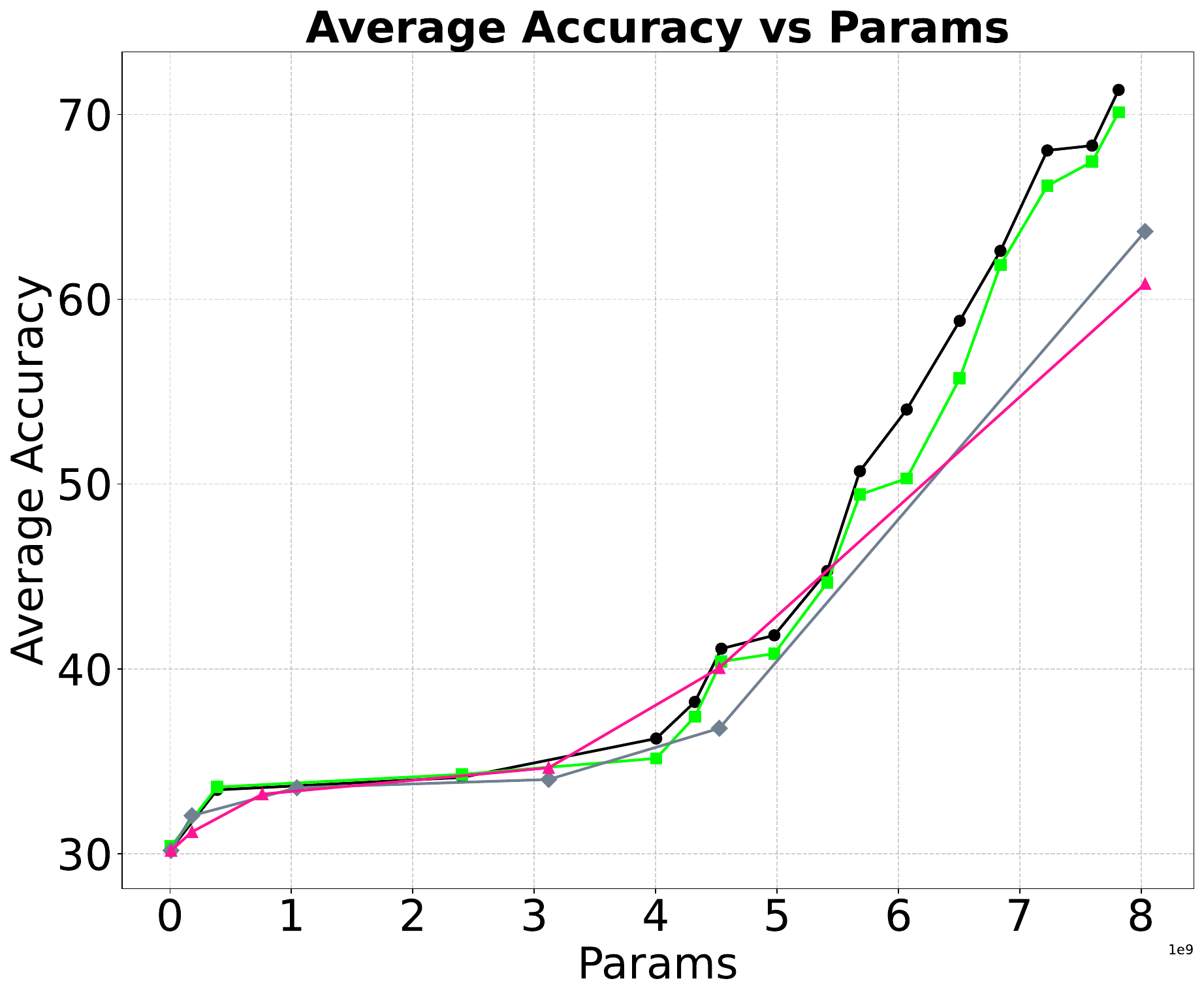}
    \includegraphics[width=.49\linewidth]{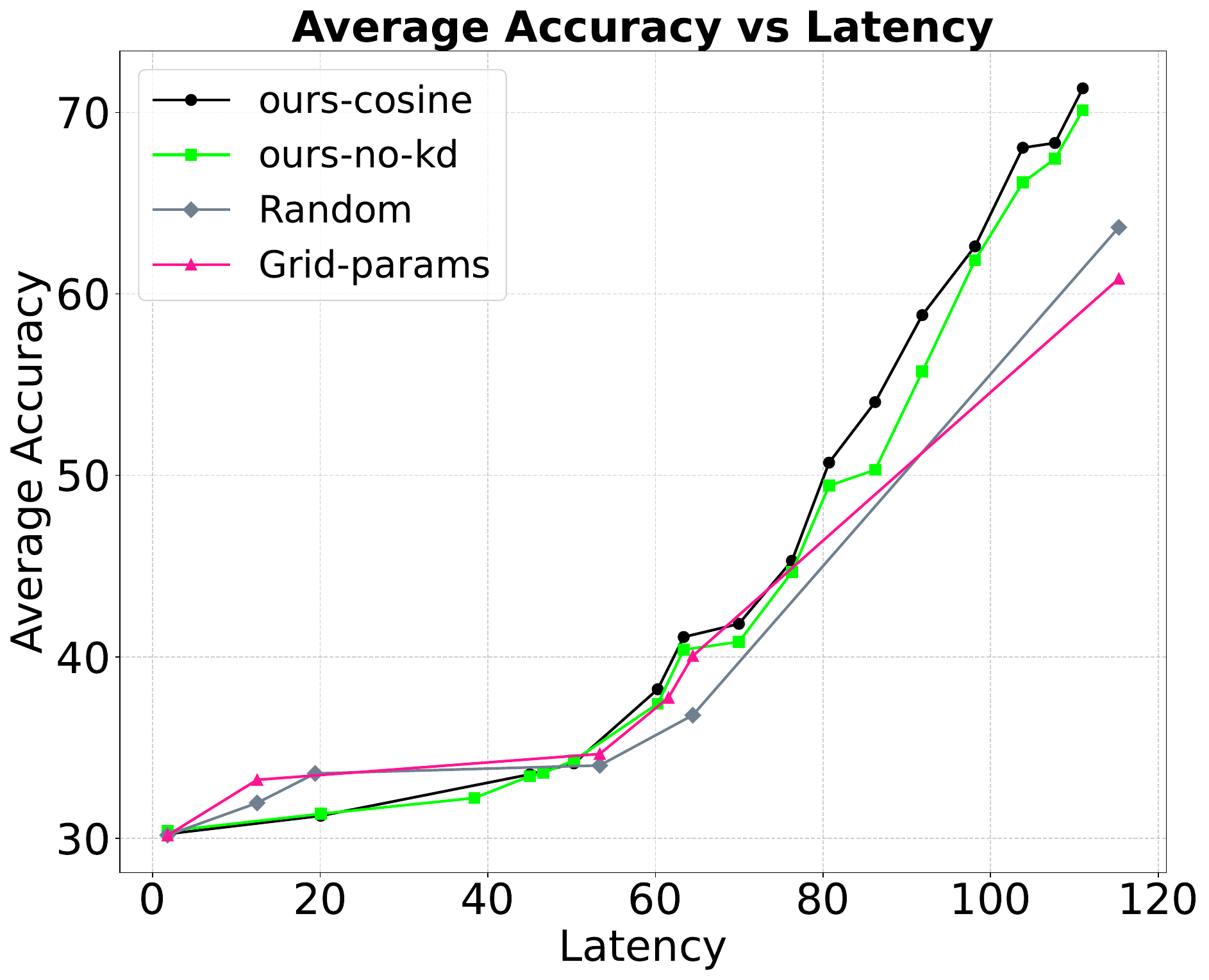}
    \caption{Comparison of Accuracy v/s Latency Pareto-Fronts for Different Architecture Sampling Scheme}
    \label{fig:sampling_schemes}
\end{figure}

In Figure \ref{fig:sampling_schemes}, we study the impact of different sampling schemes for sampling sub-networks in each update steps, including standard uniform sampling at random (\textbf{random}), using our grid described in Section~\ref{subsec:sampling} (\textbf{grid-params}) without importance scoring (see Section~\ref{subsec:importance}) and after importance sorting (with \textbf{ours-cosine} and without \textbf{ours-no-kd} knowledge distillation). We observe that specifically for larger parameter budgets our method outperform random and grid sampling schemes by a significant margin. Furthermore, \textbf{ours-cosine}, which uses the cosine-similarity loss for knowledge distillation improves over simply using the language modeling loss. Figure~\ref{fig:sampling_schemes}, highlights the value in importance sorting and calibration and the value in incorporating the knowledge distillation loss. 

\subsection{Additional Baselines}
\begin{figure}[t]
    \centering
    \includegraphics[width=.49\linewidth]{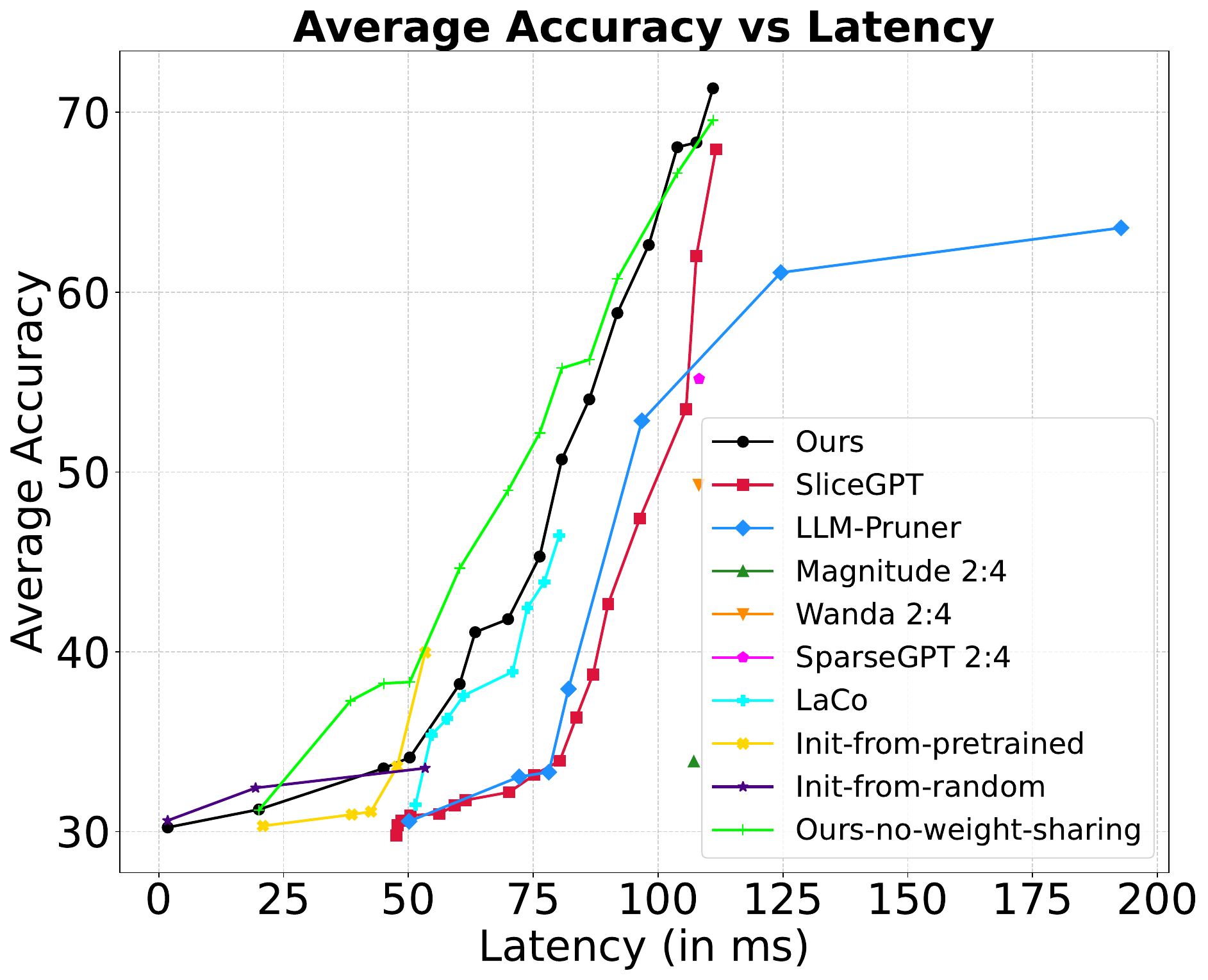}
    \includegraphics[width=.49\linewidth]{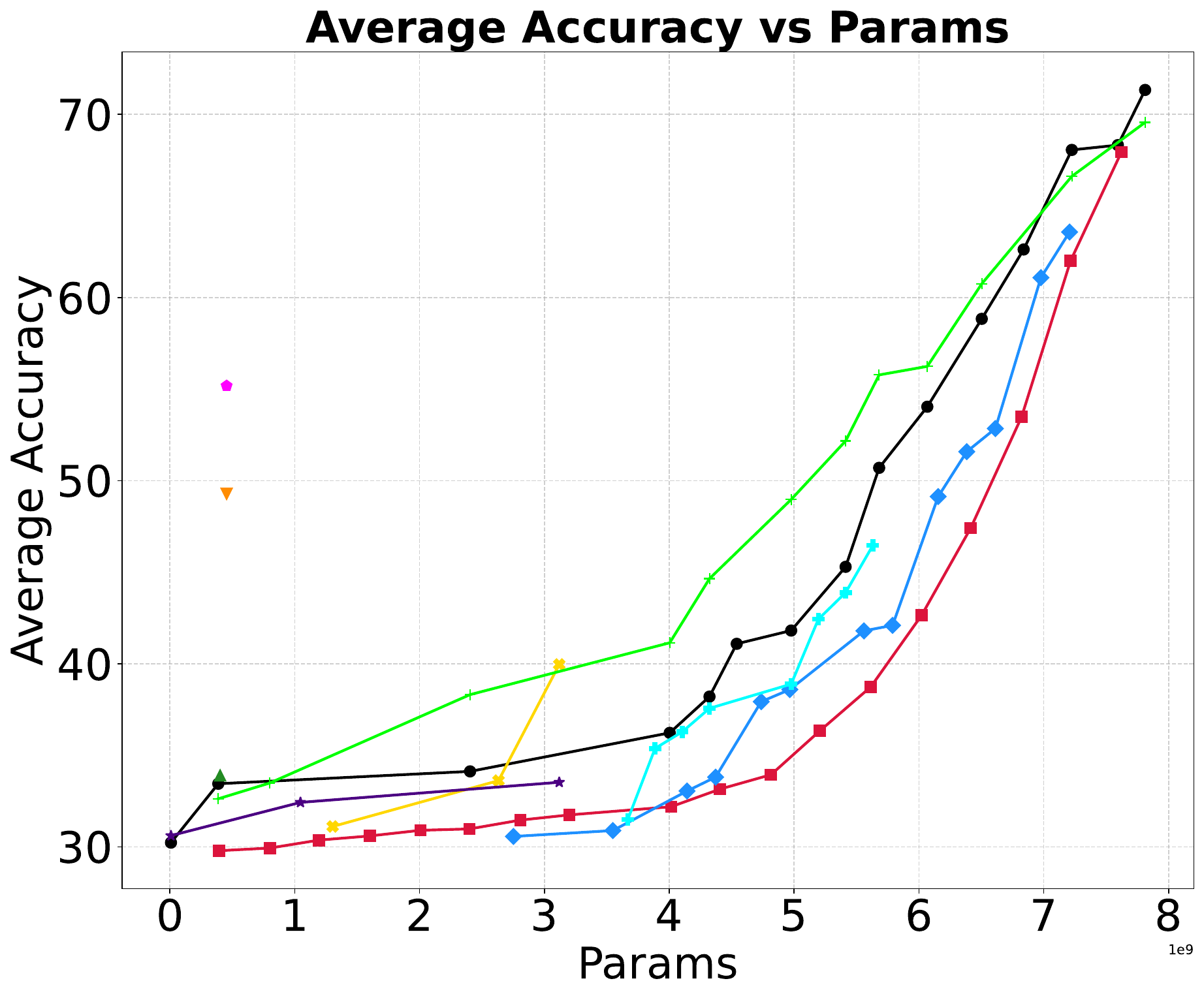}
    \caption{Comparison of Pareto-fronts for accuracy (average across commonsense reasoning tasks) v/s latency (right) and parameter count (left) for different pruning baselines and our method with and without weight sharing to compress a Llama-3.1-8B model.}
    \label{fig:baselines_extended}
\end{figure}

In addition to the baselines presented in Figure~\ref{fig:combined_plot_llama_3.1_8b}, we present a more thorough plot with all baselines presented in Table~\ref{tab:pruning-evaluation}  and our method with a higher budget \emph{ours-no-weight-sharing} included in Figure~\ref{fig:baselines_extended}.

\subsection{Comparison of Different KD Losses}\label{app:kd_loss}

\begin{table}[h]
    \centering
    \renewcommand{\arraystretch}{1.5}
    \resizebox{0.5\textwidth}{!}{\begin{tabular}{cc}
    \toprule
        \textbf{Loss Type} & \textbf{Equation} \\
        \midrule
        Forward KLD & \( \sum p(t) \log \frac{p(t)}{q(t)} \) \\
        Reverse KLD & \( \sum q(t) \log \frac{q(t)}{p(t)} \) \\
        JS Divergence & 
        \( \frac{1}{2} \left( \sum p(t) \log \frac{2p(t)}{p(t) + q(t)} + \sum q(t) \log \frac{2q(t)}{p(t) + q(t)} \right) \) \\
        L2-Norm Distance & \( \left\| \theta_T(x) - \theta_S(x) \right\|_2 \) \\
        L1-Norm Distance & \( \left\| \theta_T(x) - \theta_S(x) \right\|_1 \) \\
    \bottomrule
    \end{tabular}}
    \caption{Summary of possible forms for \(  \mathcal{D} \) in knowledge distillation. Here, \( p(t) \) and \( q(t) \) represent the teacher and student distributions, respectively, while \( \theta_{T} \) and \( \theta_{S} \) denote the teacher and student sub-networks respectively and \( \theta_{T}(x)\) and \( \theta_{S}(x)\) correspond to their respective output logits.}
    \label{tab:kd_loss_forms}
\end{table}

\begin{figure}[t]
    \centering
    \includegraphics[width=.49\linewidth]{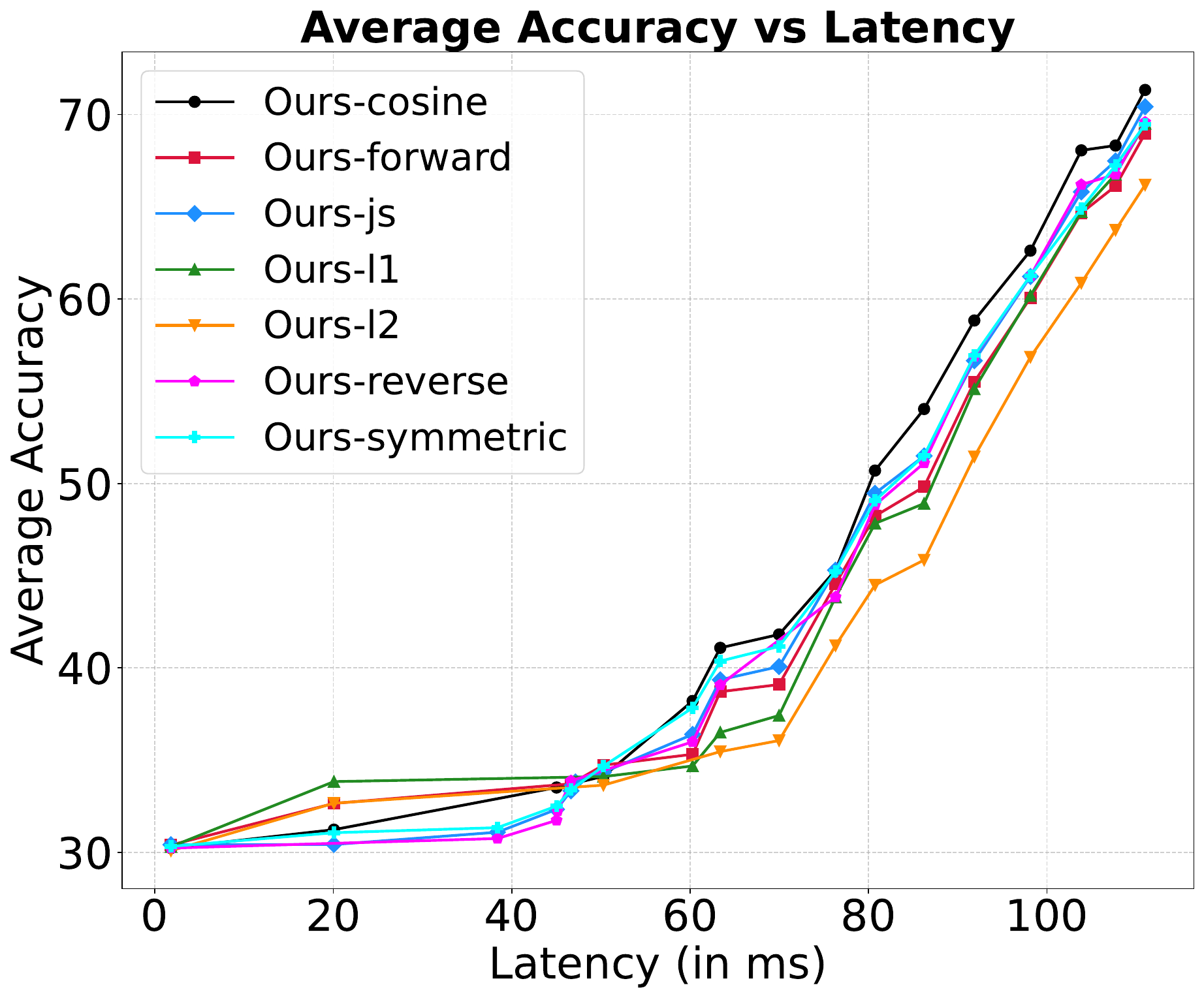}
    \includegraphics[width=.49\linewidth]{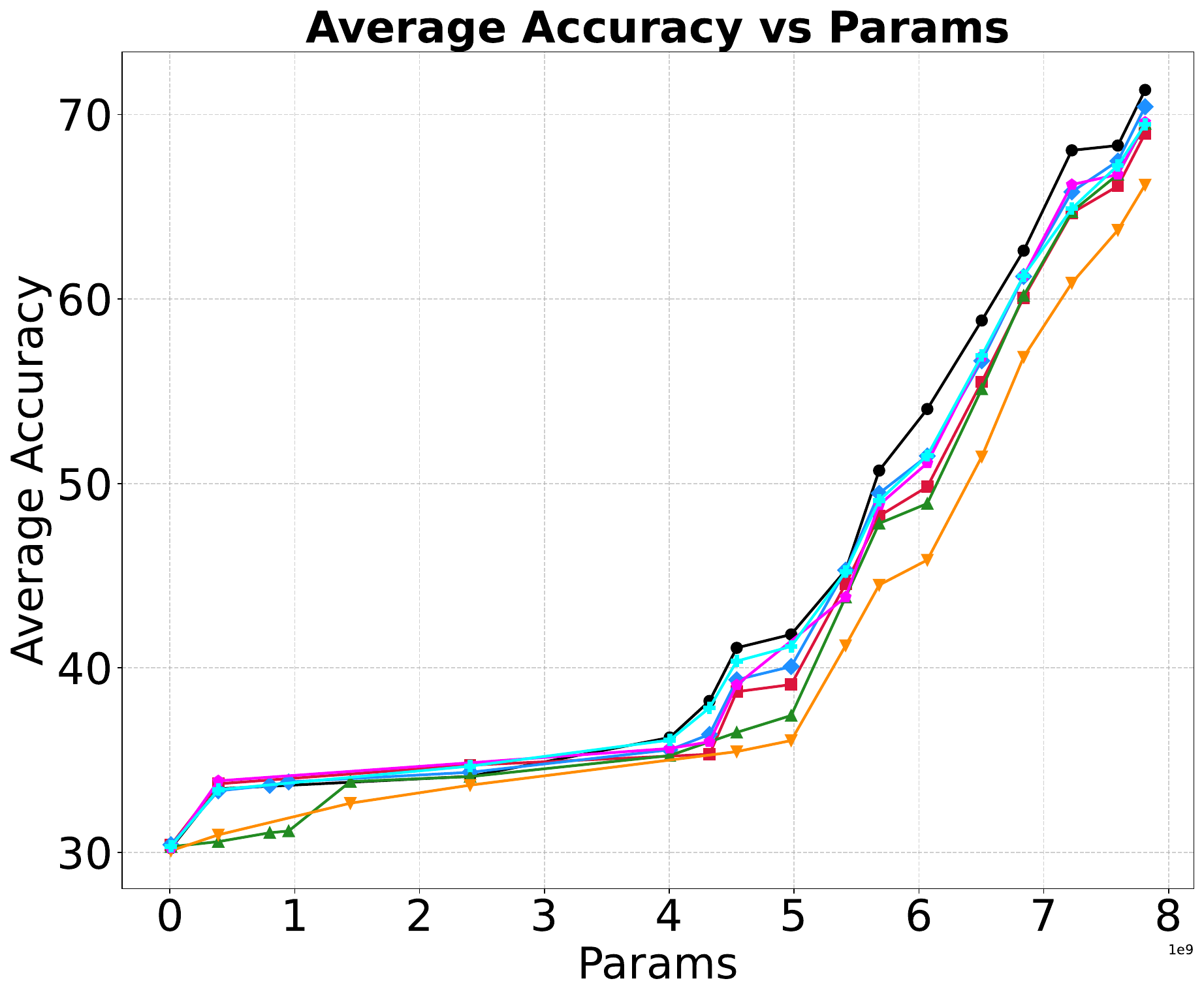}
    \caption{Comparison of Pareto-fronts for accuracy (average across commonsense reasoning tasks) v/s latency (right) and parameter count (left) for different in-place KD losses.}
    \label{fig:kd_losses}
\end{figure}

We also ablate choices of different in-place knowledge distillation loss functions to improve sub-networks shown in Table~\ref{tab:kd_loss_forms}. We observe that \emph{cosine-similarity} outperforms other KD losses in terms of the quality of sub-networks as seen in Figure \ref{fig:kd_losses}, while \emph{l2} loss performs the worst. 
We use cosine-similarity as in-place KD loss in the all experiments in the main paper.

\end{document}